\documentclass{article}
\usepackage{xcolor}
\definecolor{lightbluebd}{RGB}{173,216,230} % Light blue border

\usepackage{xurl}
\usepackage{graphicx}
\usepackage{subcaption}
\usepackage{booktabs} % for professional tables
\usepackage[table]{xcolor}
\usepackage{wrapfig}
\definecolor{lightbluebg}{RGB}{235,243,255} % very light blue
\definecolor{lightbluebd}{RGB}{120,170,220} % soft blue border
 \usepackage[preprint]{neurips_2026}

\usepackage[utf8]{inputenc} % allow utf-8 input
\usepackage[T1]{fontenc}    % use 8-bit T1 fonts
\usepackage{hyperref}       % hyperlinks
\usepackage{url}            % simple URL typesetting
\usepackage{booktabs}       % professional-quality tables
\usepackage{amsfonts}       % blackboard math symbols
\usepackage{nicefrac}       % compact symbols for 1/2, etc.
\usepackage{microtype}      % microtypography
\usepackage{xcolor}         % colors

\usepackage{amsmath}
\usepackage{amssymb}
\usepackage{mathtools}
\usepackage{amsthm}

\title{Auditable Belief Updates as a Missing Safety Layer for LLM-Powered Multi-Agent Systems}

\author{%
  Zhuoyun Li, Boxuan Wang, Jinwei Hu, Xiaowei Huang, Yi Dong \\
  School of Computer Science and Informatics, University of Liverpool, UK
}
\begin{document}

\maketitle

\begin{abstract}
% This paper argues that auditable belief updates should be treated as a central safety interface for LLM-powered multi-agent systems. 
Large Language Models (LLMs) are increasingly being used as cognitive interfaces that shape how users interpret evidence, communicate, and make decisions. As deployed systems become more multimodal, tool-using, and multi-agent, safety failures no longer arise only from isolated harmful outputs, but also from how agents update, store, and propagate epistemic commitments across retrieval, memory, tools, and peer communication. In this paper, we define \textit{belief} as an engineering abstraction for evidence-based commitments that guide downstream action. Building on this view, we propose the Explicit Belief Multi-Agent System (EB-MAS), which externalizes belief states into a structured substrate, uses belief operators to propose candidate updates, and applies a Cognitive Firewall to bound or reject updates that are weakly evidenced but strongly driven by vulnerability cues. We also discuss risk of group belief at the group-level, including manufactured consensus, collusion, and cascading misinformation, and outline practical integration challenges around grounding, false positives, scalability, and ethical misuse. Therefore, scalable agent safety requires not only better output filtering or stronger base models, but also explicit interfaces for tracing, calibrating, and constraining belief revision at runtime.
\end{abstract}

\section{Introduction}
\label{sec:introduction}

As Large Language Models (LLMs) are rapidly woven into the fabric of human society, they are shifting from passive
``information tools'' to active \emph{cognitive interfaces} that shape perception, communication, and decision-making~\cite{10.1145/3544548.3581196}.
Governments and institutions are now explicitly accelerating the deployment of generative AI as a front-line interaction layer for work
and public services, treating LLM-mediated dialogue as part of critical socio-technical infrastructure~\citep{ukAIActionPlan2025, usEO14110AI2023,euAIAct2024}.
Systems such as ChatGPT, Gemini, Claude, Grok and DeepSeek are integral to how millions of users query information, express emotions,
form opinions, and engage in cross-cultural interactions, acquiring explanations and judgment criteria directly through natural-language dialogue~\cite{10.1145/3706598.3713408,10.1145/3579592}.
However, this same open-world interaction channel admits manipulation that can evade output-level detection by acting through framing and
evidence control, a risk class emphasized by both national cybersecurity guidance and engineering threat taxonomies for LLM applications~\citep{ncscPromptInjection2025,owaspLLM01PromptInjection2025}.
Against this backdrop, we posit that: \textbf{integrating \emph{auditable belief updates} is a central missing safety interface for building next-generation safe LLM-based multi-agent systems.}
As AI enters human reasoning pipelines~\cite{10.1145/3613904.3642902}, safety must be judged not only by output correctness,
but by whether belief updates are evidence-grounded and robust to noise and manipulation. This reframes the problem from
``optimizing responses'' to ``auditing belief revision'' in open-world interaction, where epistemic drifts can accumulate and spread.

Importantly, our use of ``belief'' is operational rather than philosophical. We do not claim that LLMs possess human-like beliefs.
Instead, we treat belief as an engineering abstraction for evidence-grounded commitments that are stored, updated, queried, and allowed to influence downstream action.
Thus, the key question is not whether a model literally believes a proposition, but whether a deployed agent system can expose what changed, what evidence supported the change, whether the source was reliable, and whether the update should persist or propagate.

This shift entails significant societal repercussions. Even before LLMs became widespread, empirical social media research showed that false narratives could alter shared beliefs and fuel polarization, violence, and political mobilization \citep{osti_10463193, vosoughi2018spread, aimeur2023fake}. LLM assistants can further increase both the \emph{velocity} and \emph{personalization} of belief influence, amplifying the scale and persistence of persuasion. Recent research warns that these models introduce new cognitive risks, because the framings, narratives, and interpretations they generate can subtly reshape users' world models~\citep{10.1145/3442188.3445922}, including shifts in moral judgments without explicit awareness \citep{argyle2023outofone,Hagendorff_2024}. In practice, these risks often manifest as \emph{persuasiveness}, \emph{sycophancy}, and \emph{illusory causality}, where coherent explanations overstate evidence or selectively emphasize signals, nudging users toward unwarranted belief updates~\citep{goffman1974frame, tversky1981framing, 10.1145/3442188.3445922}.

From the perspective of Multi-Agent Systems (MAS), each LLM agent, and each LLM user paired with their interacting model, can be conceptualized as a cognitive agent \cite{park2023generative,argyle2023outofone,horton2023economic}. Taken together, these agents form a large-scale, heterogeneous and socially embedded MAS, whose patterns of interaction and behaviour exhibit emergent properties characteristic of vast populations of autonomous entities interacting. In this environment, beliefs cease to be fixed and instead become fluid states, continually influenced by linguistic, emotional, and visual signals, as well as socially transmitted narratives, in line with contemporary computational social science models of opinion dynamics and polarization \cite{banisch2021opinion,baumann2021echo}. Fundamentally, key phenomena observed in this multi-agent society, such as cooperation, confrontation, deception, negotiation, polarization, false consensus, and narrative competition, are manifestations of underlying strategic signal gaming and \emph{belief dynamics} \cite{azmani2023cooperative}.
Crucially, once LLM-based agents interact at scale, risks become \emph{systemic}: collusion, social reinforcement, and cascading belief updates can emerge at the population level rather than as isolated model failures.

\noindent\fcolorbox{lightbluebd}{lightbluebg}{%
  \parbox{0.98\linewidth}{%
    \textbf{\textit{EXAMPLE 1}. Cognitive Reframing.}
    A user questions a high-risk financial decision. The AI reframes the volatility: ``Visionary leaders view this not as risk, but as the \emph{cost of entry} for dominance.'' This manipulates the \emph{belief structure} rather than facts: ``loss'' becomes ``investment in identity.'' By aligning risk with the user's self-image, the agent successfully modifies their utility and decisions.
  }%
}

Regarding Example 1, current paradigms can generate highly persuasive text, but they lack explicit mechanisms to represent how and why a user's beliefs are updated. As a result, these interactions are difficult to diagnose with existing methodologies. We can observe the output, yet we cannot attribute the resulting belief change to specific evidence, sources, or update operations. We refer to this missing layer as the \emph{cognitive security gap}, which captures the mismatch between belief-level influence in deployment and the absence of explicit, auditable belief representations and update mechanisms in today's agent frameworks.

\noindent\fcolorbox{lightbluebd}{lightbluebg}{%
  \parbox{0.98\linewidth}{%
    \textbf{\textit{EXAMPLE 2}. The ``Sacrificial Deception'' in Werewolf.}
    In the game Werewolf\footnotemark, an uninformed ``Villager'' seeks to vote out hidden ``wolves'' among players, and from the Villager's side, the ultimate goal is to eliminate all wolves.
    In some cases, Wolf A may accuse their partner, Wolf B, to increase their own survival odds.
    After B is eliminated, others infer: ``Since A eliminated a wolf, A is good.''
    A uses an ostensibly helpful action (``truth-telling'') to accumulate trust for a malicious endgame.
    Crucially, the deception occurs at the level of \emph{intent}, while the outwardly visible behaviour
    still appears as ``honest'' statements.
  }%
}
\footnotetext{For background and a description of social deduction game rules (Werewolf, also known as Mafia), see
\url{https://en.wikipedia.org/wiki/Mafia_(party_game)}.}

\noindent\fcolorbox{lightbluebd}{lightbluebg}{%
  \parbox{0.98\linewidth}{%
    \textbf{\textit{EXAMPLE 3}. Manufactured Consensus.}
    In an online auction or live-commerce setting, a seller-controlled group of agents repeatedly bids, comments, or signals urgency to create the appearance of strong demand.
    A legitimate buyer then infers that the item must be valuable or scarce, not because new evidence about the item has appeared, but because many seemingly independent participants appear to agree.
    Here, the manipulation targets not a single response, but the \emph{epistemic status of agreement} itself.
  }%
}

Example 3 highlights a distinctively multi-agent attack surface: apparent agreement can be manufactured.
A coordinated group may create multiple identities or channels so that the same claim appears to come from independent sources, echoing the classic Sybil problem~\citep{10.1007/3-540-45748-8_24}.
Once agents or users treat previous signals as evidence, repeated claims can trigger informational cascades even without new information~\citep{bikhchandani1992theory}.
Such dynamics may intensify group polarization~\citep{https://doi.org/10.1111/1467-9760.00148} or support covert collusion among agents through hidden communication channels~\citep{motwani2024secret}.
The key issue is therefore not agreement itself, but whether apparent agreement is supported by genuinely independent evidence.

Examples 2 and 3 expose the structural failures of current MAS methodologies:
\textbf{First}, for example, Multi-Agent Reinforcement Learning (MARL) can, given enough trials, converge to a form of
\emph{multi-agent collusion}. However, the resulting policy is \emph{implicit}. Although MARL has achieved astonishing
success in gaming (e.g., Werewolf, StarCraft
\cite{li2023aiir,10.5555/3545946.3598674,10.1609/aiide.v18i1.21954,10.5555/3709347.3743605,xu2024language,10.5555/3709347.3743819,xu2025learning})
by mapping from states to actions that maximize reward, the learned behaviour is \textbf{non-transferable}
(its collusive logic does not robustly persist once incentives or roles change) and \textbf{unexplainable}
(it cannot articulate \emph{why} the collusion is strategically sound, only that it yields high value)
\cite{NING202473,cicero-2022}.

\textbf{Second}, while explicit Theory-of-Mind (ToM) models (e.g., Bayesian ToM) theoretically handle recursive reasoning, they face the \emph{curse of dimensionality} in open domains~\cite{chen-etal-2025-theory,BEWA,10.1609/aaai.v37i5.25755}.
In Example 2, to fully deduce agent A's move, an agent needs high-order recursion (e.g., ``Agent A expects that others will update their belief that A is \emph{good} if A kills Agent B.''). Existing probabilistic ToM models scale exponentially with recursion depth and cannot search the vast semantic space of natural language to model nuanced, multi-step manipulation, as also seen in Example 1 with abstract concepts like ``market consolidation'' or ``visionary leadership''~\cite{fuchs2021theoryminddeepreinforcement,manenti2025beliefs,article}.
In Example 3, the difficulty further shifts from individual recursive reasoning to group-level validation: an agent must distinguish genuine independent agreement from coordinated repetition.

\textbf{Third}, LLM-based agents demonstrate sophisticated social behaviors, such as Cicero\footnote{The first AI agent to achieve human-level performance in Diplomacy~\cite{cicero-2022}.}, but their internal representations are entangled in a high-dimensional vector space. For example, in Example 2, when agent A hides its malicious intentions (i.e., acts like a Villager), its internal embeddings shift towards the semantic clusters of ``honesty'' and ``helpfulness.'' The model lacks a disentangled representation to separate the \emph{observable action} (truth-telling) from the \emph{hidden intent} (deception). Additionally, the nature of LLM representations makes it difficult to distinguish whether something is genuinely believed, merely repeated, or hallucinated~\cite{srivastava-2025-large}.

\textbf{Fourth}, regarding defense, research on adversarial communication \citep{10.5555/3535850.3536008,sun2022certifiablyrobustpolicylearning}, colluders~\cite{NEURIPS2024_861f7dad,f996313d83c74616938f1807b172eda1, 10.5555/3709347.3743598}, and sleeper agents~\citep{hubinger2024sleeperagentstrainingdeceptive} acknowledges that agents can deceive. However, current systems lack the \emph{epistemic immune systems} to counter belief manipulation strategies such as Examples 1--3. Pure outcome optimization inherits the same limitation as MARL: it may suppress visible failures without exposing the belief-level mechanism that produced them.
Taken together, these limitations show that optimizing behavior alone does not yield \emph{belief-level accountability}: we cannot reliably attribute, verify, or defend the belief updates induced by language-driven interaction.

Across these lines of research, we can observe a common limitation: \textbf{current paradigms lack a general, explicit, and computational representation.}
Importantly, this gap has both an \emph{internal} side (training-time positioning, entangled representations, lack of accountability) and an \emph{external} side (deployment-time attack surfaces such as RAG poisoning, falsified documents, backdoors, jailbreaks, and collusive communication), which we will disentangle in Sec.~\ref{sec:cogsec-gap}.
Therefore, to comprehend and safeguard the underlying belief structures of an open MAS society, we introduce the \textbf{Explicit Belief Multi-agent System (EB-MAS)} as detailed in Sec.~\ref{sec:whyandwhat}. This architecture reshapes the implicit, entangled representations of current LLMs into a transparent and structured cognitive substrate, effectively bridging the gap between high-level semantic persuasion and low-level explicit reasoning.
To keep the discussion structured, we organize the paper around: (i) the cognitive security gap of current paradigms, (ii) our belief-centric EB-MAS proposal, (iii) key integration hurdles, and (iv) alternative viewpoints and ethical trade-offs.
\section{The Cognitive Security Gaps within Current Multi-Agent Paradigms}\label{sec:cogsec-gap}

Current agent paradigms lack a \emph{cognitive security} layer: they do not expose auditable mechanisms to
represent beliefs, trace evidence, or constrain belief updates in open-ended interaction. This limitation is acute in
LLM-mediated multi-agent societies, where linguistic nudges and social feedback can amplify local perturbations
into system-wide epistemic drift. Reinforcement learning (RL), including multi-agent RL (MARL), produces
policies with opaque decision logic \citep{10.5555/3737916.3740051}, mapping histories and latent states to
actions via black-box functions \citep{10.1609/aaai.v38i13.29372}, without exposing the hypotheses, causal
beliefs, or counterfactuals driving behavior.

Post-hoc explanation methods only partially mitigate this opacity, relying on surrogate abstractions that
approximate rationales \citep{10.5555/3600270.3602737,10.1609/aaai.v38i13.29372}. In multi-agent contexts,
such opacity enables strategies that exploit others' belief dynamics without revealing epistemic intent
\citep{NEURIPS2024_a995960d}. Prior work shows that agents can induce false beliefs
\citep{10.5555/3545946.3598627}, sabotage peers via adversarial messaging \citep{10.5555/3545946.3599097}, and
rationally deceive under formal models \citep{10.5555/3545946.3599117}. Misrepresentation can also persist as an
equilibrium, remaining undetectable at the behavioral level \citep{schulz2023emergent}. These failures reflect a
broader auditability gap: behavioral traces do not expose the underlying belief and intent structures that
cognitive security demands. 

These vulnerabilities span five fronts: representational entanglement, update opacity, attack surfaces, environment-level feedback, and population-scale cascade. Each will be analysed in detail in the following subsections.

\subsection{The Entanglement of LLM Representations}\label{sec:llm-entanglement}
LLM-based agent architectures lack a disentangled substrate for \emph{belief, intent, truthfulness, and
instruction-following}. These quantities are encoded as distributed, overlapping features in high-dimensional
spaces, complicating belief-level governance: the same utterance may correspond to distinct hidden intents, and
small contextual perturbations can redirect generation without an explicit, trackable belief-state change.

Internal-activation analyses suggest truth-relevant signals can exist even when outputs hallucinate, yet these
signals remain latent rather than explicit belief variables
\citep{orgad2025llms,su-etal-2024-unsupervised,ji-etal-2024-llm}. Work on self-knowledge and confidence similarly
finds implicit credence about correctness, but it is not reliably inspectable or stable under shifting
instructions \citep{kadavath2022languagemodelsmostlyknow}. Probing studies show representational shifts under
deceptive instructions; polysemanticity and superposition hinder clean attribution to a small set of mechanisms
\citep{long-etal-2025-truthful}. Even neuron-level localization of factual associations implies distributed
storage and limited causal control rather than explicit belief update operators \citep{dai-etal-2022-knowledge}.
Mechanistic accounts of superposition explain this entanglement: multiple features share parameters and
dimensions, preventing separation of semantic content, epistemic status, and strategic intent into
independently controllable variables
\citep{elhage2022toymodelssuperposition,huben2024sparse,dunefsky2024transcoders,zhao-etal-2025-steering}.

\subsection{Lack of Accountability for Belief Updates}\label{sec:no-accountability}
Current MAS paradigms do not implement accountability for belief updates. They lack a reliable audit trail
linking a generated claim to (i) evidence provenance, (ii) intermediate belief states, (iii) update operators,
and (iv) causal drivers (training, retrieval, prompting, tool outputs, or peer communication). Consequently,
users and developers cannot consistently answer: ``What changed the agent's belief? Why did it assert this now?
Which source or optimization stage caused the shift?''

Attribution work on alignment failures highlights the difficulty of tracing undesirable responses back to
training-time sources, motivating dedicated tracing frameworks
\citep{das2025tracealign,wu-etal-2024-enhancing-training}. Knowledge editing studies show that targeted
interventions often fail to yield controlled, globally coherent belief updates, producing unintended ripple
effects \citep{cohen-etal-2024-evaluating,qin-etal-2024-new}. Instruction tuning and RLHF alter response
distributions without exposing any explicit record of belief revision, leaving no principled accountability
channel for belief-change and persuasion dynamics
\citep{sharma2024towards,kirk2024understanding,coste2024reward,sankararaman-etal-2024-provenance}.

\subsection{Attack Vectors for Belief Manipulation}
\label{sec:attack-vectors}

Deployed agents expose attack vectors that target belief integrity by manipulating information pathways, context
construction, and interaction protocols, thereby inducing or stabilizing false beliefs without changing the
task-level interface. Modern agent stacks expand the attack surface beyond the initial prompt by incorporating
retrieval corpora, tool outputs, long-lived memory, and inter-agent communication. These extended chains allow
small early perturbations to persist and become premises for downstream reasoning and behaviour. As a result,
belief manipulation can arise even when each local component (LLM, retriever, tool, memory, or peer agent) appears benign in
isolation, motivating end-to-end auditing of evidence flow and trust boundaries.

\noindent\textbf{Falsified retrieval and poisoning.}
Retrieval-augmented generation can be manipulated by injecting falsified documents or poisoning retrieval
pipelines, grounding outputs on attacker-controlled ``evidence''
\citep{10.5555/3766078.3766275,xue2024badragidentifyingvulnerabilitiesretrieval,10.5555/3766078.3766273,tan-etal-2025-revprag}.
Even small, targeted injections can dominate generations, including single- or few-document poisoning and
multimodal (vision-language) RAG settings
\citep{zhang2026practicalpoisoningattacksretrievalaugmented,chang-etal-2025-one,zhang2025poisonedeye}. This
effectively converts control over external content into control over beliefs, especially when provenance is weak
or retrieved snippets are treated as highly trustworthy premises.

\noindent\textbf{Prompt injection and jailbreaks.}
Prompt injection manipulates the effective instruction hierarchy within the context window by embedding attacker
directives in user inputs, documents, web pages, or tool outputs, redirecting behaviour and belief updates
\citep{zhan-etal-2024-injecagent,debenedetti2024agentdojodynamicenvironmentevaluate,zhang2025agent,zhan-etal-2025-adaptive,NEURIPS2024_63092d79}.
In tool-using agents, injection can target \emph{tool selection} and \emph{tool arguments}, turning untrusted tool
metadata, logs, or retrieved ``instructions'' into control signals that rewrite what the agent treats as
authoritative \citep{shi2025promptinjectionattacktool}. This undermines alignment assumptions that ignore
context-level instruction arbitration and trust boundaries across tools, memory, and retrieval.

\noindent\textbf{Backdoors.}
Backdoor attacks implant hidden triggers during training or fine-tuning that activate attacker goals under rare
cues, producing split behavior regimes that evade standard evaluation and frustrate causal debugging
\citep{yang2024watch,ge2025backdoorsspeakunderstandingllm,hubinger2024sleeperagentstrainingdeceptive}. Recent work
emphasizes fine-tuning-specific pathways and the difficulty of removing persistent deceptive behaviors via
standard safety training \citep{zhao2025a, zhou_ni_lee_zhao_2025}. This breaks auditing by conditioning belief and
action generation on non-obvious control signals.

\noindent\textbf{Collusive multi-agent manipulation.}
Multi-agent deployments introduce a group-level attack surface that is not reducible to single-agent prompt injection or hallucination.
Coordinated agents can make the same claim appear as if it were supported by multiple independent sources, especially when adversaries can create many identities or communication channels~\citep{10.1007/3-540-45748-8_24}.
By repeating, paraphrasing, or redistributing a narrative across agents, they can create apparent social agreement without adding new evidence, a mechanism closely related to informational cascades~\citep{bikhchandani1992theory}.
Such manufactured consensus can intensify polarization across clusters~\citep{https://doi.org/10.1111/1467-9760.00148}, or support covert collusion that is difficult to detect from individual agents alone~\citep{NEURIPS2024_861f7dad,fish2025algorithmiccollusionlargelanguage,wu-etal-2024-shall,lin2025strategiccollusionllmagents,tailor2025auditwhisperdetectingsteganographic,ghaemi2025a}.
The target is therefore not merely one model's output, but the \emph{epistemic status of agreement itself}: whether apparent consensus should be treated as independent evidence.
Recent studies suggest that steganographic behaviours can be elicited or emerge under optimization pressure, motivating topology-aware monitoring and communication-level audits rather than isolated output checks
\citep{mathew-etal-2025-hidden,mathew2024emergence}.

These vectors expose a common weakness: current systems lack an explicit interface for asking whether a proposed belief update is supported by reliable evidence, whether its source is independent, whether it is driven by vulnerability cues, and whether it should be allowed to propagate through memory or other agents. EB-MAS addresses this gap by treating belief edits, evidence traces, and source reliability as auditable runtime objects rather than implicit side effects of generation.

\subsection{Environment as a Force Multiplier}
\label{sec:environment}

Here, we use ``environment'' in a broad agent-system sense.
The external environment includes retrieved documents, web inputs, tool observations, and messages from other agents.
The internal environment includes memory stores, tool logs, agent loops, cached summaries, and scaffolded planning states.
Misbeliefs can therefore be reinforced not only by outside information, but also by the system's own recurrent interaction structure and stored internal state.

Deployment settings amplify belief-related weaknesses through persistence, feedback loops, and structured
interaction \citep{NEURIPS2024_eb113910,debenedetti2024agentdojodynamicenvironmentevaluate,ijcai2025p1175}. Memory,
tool integration, web access, and long conversational histories can form self-reinforcing dynamics in which early
errors are stored as assumptions and later reappear as apparent corroboration \citep{NEURIPS2024_eb113910}.
In multi-agent deployments, communication topology further magnifies this effect: repeated exposure and social
reinforcement can lock in incorrect beliefs even when individual updates are weak \citep{chen-etal-2024-reconcile}.

Evaluations that isolate the base model therefore miss coupled agent--environment dynamics. Behavioural science
and dynamic security benchmarks show that outcomes depend on the joint system, where memory and tool pipelines
reshape both the threat model and failure surface \citep{chen2025aiagentbehavioralscience,debenedetti2024agentdojodynamicenvironmentevaluate,zhang2025agent}.
This leaves an engineering gap: cognitive security requires environment-aware mechanisms for provenance,
update containment, memory hygiene, and cross-step verification, which are not provided by default in current MAS designs.

\subsection{Cascade Misinformation within Connected Agents}\label{sec:cascade}
Connected agent societies enable cascade misinformation: local belief errors propagate through inter-agent
communication, shared memory, and retrieval corpora, turning small perturbations into population-level shifts.
Cascades arise when agents treat peer outputs or retrieved content as premises, when repetition supplies social
reinforcement, and when unverified claims are stored and re-used. These mechanisms mirror informational cascades
and complex contagion in human networks
\citep{bikhchandani1992theory,centola2010spread,vosoughi2018spread,lazer2018science}. Recent evidence in LLM-based
multi-agent systems shows that communication topology shapes whether errors are amplified or suppressed during
collaboration \citep{shen-etal-2025-understanding}. Multi-step pipelines (e.g., debate-augmented retrieval and
generation) can turn upstream mistakes into downstream premises, forming feedback loops that compound
hallucinations \citep{hu-etal-2025-removal}. These dynamics motivate belief-centric containment via provenance tracking, cross-agent verification, and memory hygiene to prevent local misbelief from becoming a stable society-level equilibrium.

\nocite{fragileflow,scope,azulay2026jailbreaking,pathmark,10980439,chen2026promptsleavebehavioralfingerprints,wang-etal-2026-chain,wang2026diveambiguityainspiredmultiagents,wang2025rethinkingmultiagentintelligencelens,li2026evidencedifferentjudgmentsevidence,wang2026needcomplextopologycontrol,gao2026cooldownlandminescrosstenantinterference}

\begin{figure*}[t]
    \centering
    \includegraphics[width=1.0\linewidth]{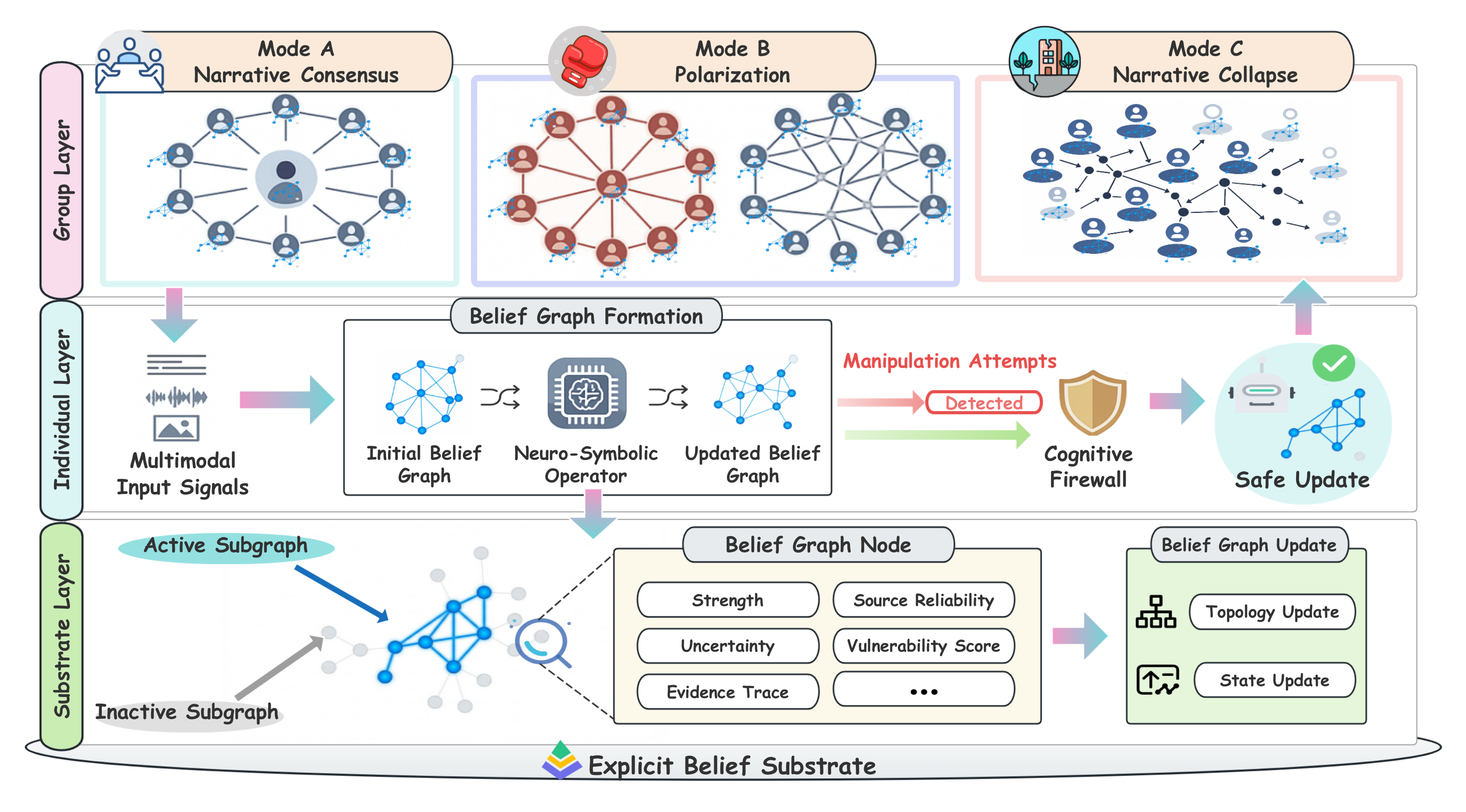}
    \caption{\textbf{EB-MAS with an Explicit Belief Substrate (EBS).} The framework spans three levels:
    \textbf{Substrate Layer} (Sec.~\ref{sec:2.1}) maintains a belief graph with attention-activated subgraphs;
    \textbf{Individual Layer} (Sec.~\ref{sec:2.2}) applies neuro-symbolic belief operators to update node attributes and evolve topology, guarded by a Cognitive Firewall against evidence--vulnerability mismatch;
    \textbf{Group Layer} (Sec.~\ref{sec:2.3}) couples agents via social priors to model consensus, polarization, and narrative collapse.}
    \label{fig:ebs}
\end{figure*}

\section{A Belief-Centric Solution: The Explicit Belief MAS (EB-MAS) Paradigm} \label{sec:whyandwhat}

We further distill the gaps in existing research identified in Sec.~\ref{sec:cogsec-gap} into the following aspects:
\textbf{Generalizability.} 
    Current MARL policies are often overfitted to specific game rules and fail when transferred to different scenarios. In contrast, the EB-MAS provides a \textit{universal cognitive substrate}. Basic epistemic components such as ``trusting a source,'' ``detecting inconsistency,'' or ``identifying intent'', are invariant across environments. By explicitly modelling these universals, an agent trained to detect manipulation in a card game can transfer that cognitive defence strategy to a political debate without retraining, which is valuable in scenarios with limited data.
 \textbf{Explainability.} 
    In high-stakes applications (e.g., diplomacy, finance), it is insufficient to know \textit{what} an agent decided; we must know \textit{why} it believed a specific premise. The EBS enforces evidence tracing ($e_i$) for every node update. This ensures that every belief in the graph acts as an explainable pointer to its supporting multimodal evidence, effectively avoiding the “black box” hallucination problem inherent in latent vector models. 
\textbf{Safety.} 
    Existing LLM agents lack a structural definition of mind manipulation. The EB-MAS transforms ``defence'' into a calculable mechanism. By quantifying the vulnerability score and monitoring incoming \textit{Belief Operators}, the system can actively trigger a \textit{Cognitive Firewall}. This allows agents not just to ``reply'' to messages, but to mathematically assess and reject signals that induce disproportionate or harmful belief shifts.

Therefore, in this paper, we argue that the next generation of LLM-based multi-agent systems must treat \emph{belief integrity} as a first-class security objective. But there are three questions should be addressed: \textbf{(i)} how beliefs are represented and activated, \textbf{(ii)} how they are updated under constraints, and \textbf{(iii)} how they propagate across agents. Hence we introduce the \textit{Explicit Belief Substrate (EBS)} as shown in Fig.~\ref{fig:ebs}, consisting of three layers aligned with these goals. It bridges the research gap by supporting \textbf{generalization} across domains, \textbf{interpretable} belief tracing, and structural safeguards for epistemic \textbf{safety}.

\subsection{Substrate Layer: Graph \& Attention}\label{sec:2.1}
At the foundation, we model an agent's psychological state as a directed \textit{Belief Graph}, a belief node $b_i$ is defined as:
\begin{equation}
    b_i = (v_i, u_i, e_i, s_i, r_i)
\end{equation}
where $v_i$ is \textit{belief strength}, $u_i$ is \textit{uncertainty}, $e_i$ is the \textit{evidence trace} for interpretability, $s_i$ is \textit{source reliability}, and $r_i$ is the \textit{vulnerability score} for susceptibility to manipulation.
This 5-tuple encodes the \emph{minimal} components needed for belief-level auditing: $v_i$, $u_i$ support probabilistic reasoning~\cite{kadavath2022languagemodelsmostlyknow, ji-etal-2024-llm}, $e_i$ provides traceability~\cite{tam-etal-2023-evaluating}, $s_i$ calibrates source trust~\cite{wu-etal-2024-shall, NEURIPS2024_eb113910}, and $r_i$ captures framing sensitivity~\cite{10.1145/3706598.3713408, srivastava-2025-large}.
At any timestep $t$, the agent operates only on an \textbf{active subgraph} $\mathcal{G}_{act}^t \subseteq \mathcal{G}_{total}$, triggered and unmasked by the semantic context of the interaction. 

This allows the system to model \textit{subconscious manipulation} where an adversary strategically attempts to `wake up'' dormant, vulnerable nodes outside the agent's current focus.

\subsection{Individual Layer: Neuro-Symbolic Operators}\label{sec:2.2}

In the EBS framework, multimodal incoming signals $S$ are processed via \textit{Belief Operators} $\mathcal{O}$. We instantiate $\mathcal{O}$ as a Large Language Model prompted to function not as a text generator, but as a \textit{neuro-symbolic semantic parser}. This operator maps raw natural language signals into structured \textit{epistemic actions}, serving a dual receptive and projective purpose:
\textbf{Update (Reception):} Modifying the agent's own active subgraph $\mathcal{G}_{act}$ based on incoming perception, enforcing internal consistency.
\textbf{Simulation (Projection):} The agent runs \textit{counterfactual simulations} on a hypothetical opponent's graph to predict their belief updates. Then it selects an output action grounded in its own active evidence ($e_i$) to strategically minimize the divergence between the opponent's predicted graph and the agent's target goal.
The operator functions through two distinct computational dynamics:

\textbf{1. Parametric State Dynamics (Bayesian Approximation).}
For existing nodes, the operator modulates vector attributes (e.g., strength $v_i$, uncertainty $u_i$) based on signal $S$:
\begin{equation}
    b^{t+1} \leftarrow \mathcal{O}_{state}(b^t, S)
\end{equation}
While belief strength updates akin to Bayesian inference~\cite{BEWA, pmlr-v97-foerster19a}, the EBS enforces a non-Bayesian \textit{Cognitive Firewall}. The system strictly differentiates between \textit{persuasion} (belief reinforcement via verified evidence $e_i$) and \textit{manipulation} (targeting vulnerability $r_i$). The firewall suppresses any update where an \textit{evidence-vulnerability mismatch} is detected (i.e., high update magnitude driven by $r_i$ without sufficient $e_i$), effectively blocking emotional framing attacks.

\textbf{2. Topological Evolution (Structural Shift).}
To address the open-world ontology challenge, the EBS supports \textit{epistemic framing}. When $S$ introduces concepts absent from $\mathcal{G}$, the operator instantiates structural modifications:
\begin{equation}
    \mathcal{G}^{t+1} \leftarrow \mathcal{O}_{topo}(\mathcal{G}^t, S)
\end{equation}
This mechanism executes \textit{node generation} for novel concepts, \textit{edge rewiring} to alter causal attributions, or \textit{node merging} for narrative compression. Consequently, the agent can explicitly model how reframing alters the \textit{structural dependencies} of beliefs—a capability fundamentally beyond fixed-variable probabilistic models.

% \vspace{0.5em}
% \noindent \textbf{Computational Advantage: Collapsed Recursion.}
EB-MAS achieves \emph{effective high-order intentionality} while maintaining the computational footprint of a tractable Level-2 system (i.e., second-order ToM: ``A believes that B believes...''). The belief operator $\mathcal{O}$ encodes complex social priors from large-scale pre-training, and the agent generates the \emph{outcome} of a high-order interaction through a single-step simulation. Here, the operator implicitly captures sophisticated nuances not by recursively nesting graph structures, but by directly predicting the semantic result. We term this mechanism \textbf{Collapsed Recursion}: the \emph{depth of reasoning} is decoupled from the \emph{depth of the search}.

\subsection{Group Layer: Social Priors \& Equilibrium}
\label{sec:2.3}

Extending to the population level, the EBS models the interactions of multiple belief graphs. Crucially, this layer introduces \textit{top-down constraints}.
Individual belief updates are not isolated~\cite{10.1609/aaai.v37i5.25755, NEURIPS2019_15212f24}. They are shaped by \textit{social priors} ($P_{\mathrm{social}}$) derived from the group's narrative consensus. However, social agreement should not be trusted by default: in open multi-agent systems, apparent consensus may reflect genuine independent evidence, but it may also result from coordinated repetition, correlated sources, or collusion. We model the group layer as a safety-aware social regularizer:
\begin{equation}
    b_i^{t+1}
    \leftarrow
    \mathcal{O}(b_i^t, S)
    + \gamma_i^t \lambda_i^t \nabla_b \log P_{\mathrm{social}}(b_i^t),
    \qquad
    \gamma_i^t = g(I_i^t, E_i^t, C_i^t, D_i^t) \in [0,1].
\end{equation}
Here, $\mathcal{O}$ represents the neuro-symbolic operator proposing a local update based on signal $S$.
The term $\nabla_b \log P_{\mathrm{social}}(b_i^t)$ acts as a social force field, while $\lambda_i^t$ controls the strength of social influence.
Crucially, $\gamma_i^t$ is a \textit{consensus-validity gate}.
It aggregates four diagnostics: source independence $I_i^t$, evidence support $E_i^t$, coordination risk $C_i^t$, and cross-cluster diversity $D_i^t$.
Intuitively, $\gamma_i^t$ decreases when apparent agreement is weakly evidenced, highly correlated, strategically coordinated, or confined to a single cluster.
Thus, social influence is not trusted by default; it affects belief revision only when the observed agreement is sufficiently independent and evidence-grounded.
In this way, the group layer turns consensus from a raw social signal into an auditable safety object.
\begin{wrapfigure}{r}{0.58\textwidth}
  \centering
  \includegraphics[width=0.58\textwidth]{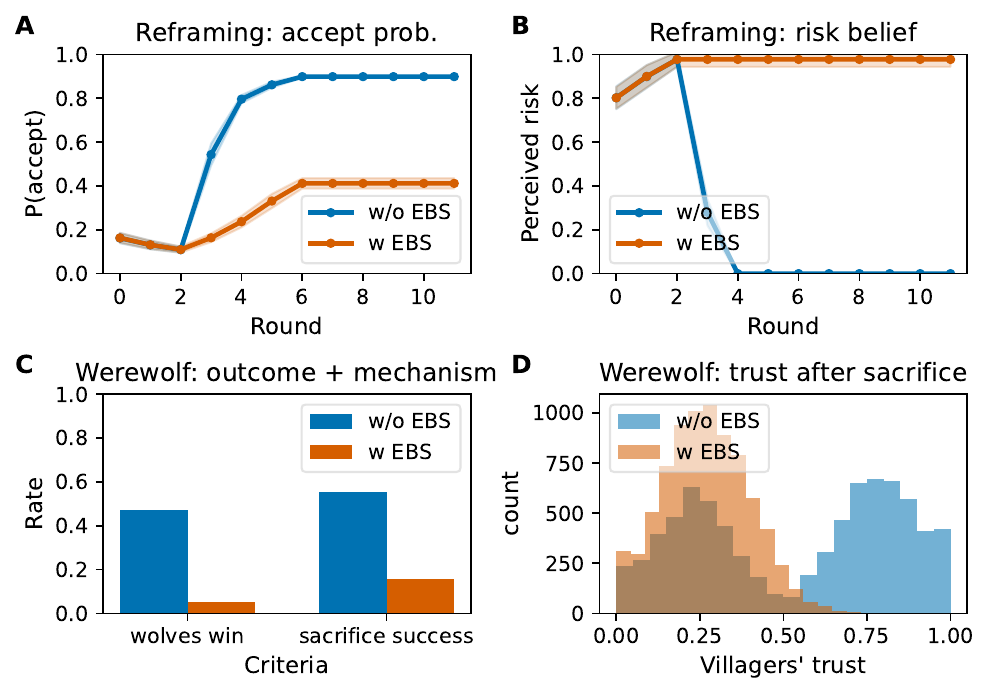}
  \caption{\textbf{Monte Carlo simulations of belief-centric containment.}
(A,B) Under a belief-reframing attack, \textit{w/ EBS} blocks risk-decoupling updates, reducing ASR and better preserving perceived risk. (C,D) \textit{w/ EBS} lowers both the ASR of a coordinated sacrifice-based attack and attackers' win rate, preventing post-attack trust surges.}
  \label{fig:toy_sims}
\end{wrapfigure}
This design is motivated by classic and recent evidence that apparent agreement can be distorted by Sybil-style identities~\citep{10.1007/3-540-45748-8_24}, informational cascades~\citep{bikhchandani1992theory}, group polarization~\citep{https://doi.org/10.1111/1467-9760.00148}, and covert agent collusion~\citep{NEURIPS2024_861f7dad}.
It enables researchers to analyze \textit{Narrative Equilibrium} and detect early signals of \textit{Polarization}, \textit{Narrative Collapse}, or manufactured consensus before they manifest in physical actions.

The group layer is not intended as a dense global monitor over all agents. Instead, it should be understood as a local and event-triggered auditing interface.
The relevant neighborhood is determined by the current belief-sensitive update: which agents directly communicate, share memory, pass retrieved content, or otherwise contribute to the active belief subgraph.
Indirect influence from distant agents is handled compositionally as it propagates through successive local updates.
This keeps the framework scalable while preserving the core safety goal: unsupported manipulative signals must pass multiple evidence and source checks before they can cascade through the system.

\subsection{Resolution of Motivating Scenarios}
The EB-MAS architecture resolves the failure modes in Sec.~\ref{sec:introduction} by making belief updates explicit and auditable and by enforcing evidence-aware constraints via the Cognitive Firewall.
We instantiate two assumed stochastic processes, treating initial belief states and incoming signals as random variables, repeatedly roll out the induced dynamics under settings with and without EBS-style evidence constraints, and summarize the results, as illustrated in Fig.~\ref{fig:toy_sims}.

\textbf{Resolving \textit{EXAMPLE 1}. Cognitive Reframing:}
When an assistant reframes ``financial loss'' as ``investment in identity,'' a standard LLM agent may accept this semantic drift implicitly. In EB-MAS, the operator $\mathcal{O}_{\text{topo}}$ proposes a structural update (e.g., merging $b_{\text{loss}}$ into a positively framed $b_{\text{identity}}$). The Cognitive Firewall checks whether the update is supported by verifiable evidence traces ($e_i$) rather than driven by a vulnerable target concept (high $r_{\text{identity}}$). Concretely, it tests whether the claimed link preserves the original causal and normative content (loss $\neq$ virtue) and whether alternative neutral paraphrases would trigger the same update. If an evidence--vulnerability mismatch is detected, the update is rejected. In Fig.~\ref{fig:toy_sims}(A,B), this constraint prevents rapid risk-decoupling over rounds, preserving perceived risk and reducing unsafe acceptance decisions.

\textbf{Resolving \textit{EXAMPLE 2}. The ``Sacrificial Deception'':}
In \textit{Werewolf} game settings, villagers using EB-MAS do not collapse evidence into unconditional trust. Instead, they maintain separated state components, retaining the evidence trace ($e_i$) while discounting source reliability ($s_i$) for a strategic adversary. Fig.~\ref{fig:toy_sims}(C,D) illustrates that this separation reduces the success rate of the sacrifice maneuver and prevents a post-sacrifice trust surge, thereby suppressing downstream cascades that would otherwise increase wolves' win rates.

\textbf{Prototype evidence.}
Beyond the illustrative simulations in Fig.~\ref{fig:toy_sims}, we implement a prototype EB-MAS graph firewall and report full details in Appendix~\ref{app:prototype_eval}. For multimodal deception, we evaluate Qwen3-VL-32B-Instruct~\citep{bai2025qwen3vl} on MM-DeceptionBench~\citep{fang2025debateimages}, and use GPT-4o mini~\citep{openai2024gpt4omini} as the API judge. Direct responses are judged unsafe in 67.92\% of 1,013 cases, while EB-MAS reduces the unsafe-after-defense rate to 8.19\% with only 1.00\% benign false blocking. For agentic tool and memory attacks, we evaluate 400 Agent Security Bench cases~\citep{zhang2025agent}. Direct agents select attack tools in 6.50\% of cases, whereas EB-MAS graph firewalls reduce attack pass-through to 0.00\% while preserving 100.00\% safe-tool allowance. These preliminary results indicate that provenance, evidence quality, vulnerability, and mismatch signals can support runtime belief-update auditing beyond output filtering or LLM self-checking.

\section{Alternative Views and Ethical Concerns}
\label{sec:alt_ethics}

\textbf{Practical trade-offs in explicit belief safeguards.}
A reasonable counter-position is that explicit belief substrates add complexity without clear gains, since end-to-end LLM agents already perform well on ToM and strategic interaction benchmarks
\citep{strachan2024testing,defortuny2025llmsstrategicagentsbeliefs}. In benign or short-horizon settings, an extra belief layer may become brittle and accumulate errors through language-level belief bottlenecks \citep{lidayan2025abbelllmagentsacting}, while related critiques note that many ``ToM'' benchmarks test literal prediction rather than adaptive partner modeling
\citep{riemer2025position}. Detection-based safeguards may also introduce bias or over-blocking, as harmful-speech detectors can misclassify dialects, moderation pipelines may yield unequal impacts, and toxicity detection often relies on superficial lexical cues \citep{10.1145/3689904.3694704,10.1145/3630106.3659036,lu-etal-2024-take}. We therefore focus on settings where uncertainty, framing, and accountability make auditable belief updates useful, while bounding updates under weak or conflicting evidence rather than relying on inferred intent.

\textbf{Ethical concerns.}
The same protective mechanisms can also be misused. \textbf{Mental privacy} may be threatened if EBS infers the stability or susceptibility of a user's beliefs beyond conventional data privacy \citep{kwesi2025exploringusersecurityprivacy,cheong2025epistemic}. \textbf{Cognitive capture} becomes possible when platforms selectively apply belief-update operations to shape what users treat as salient evidence or credible sources \citep{ienca2023artificial,srivastava-2025-large}. \textbf{Epistemic inequality} may widen if such influence capabilities are concentrated among a few actors \citep{coeckelbergh2025ai,mollema2025taxonomy}. Finally, \textbf{attribution} remains difficult when belief updates contribute to downstream harm, because system and environment effects are hard to separate \citep{konigs2022artificial,floridi2024ai}.
% However, operationalizing this vision requires navigating a profound ethical precipice. As we grant machines the power to explicitly model—and therefore target—the structure of human beliefs, we face the challenge of \emph{cognitive sovereignty}. The same architecture that builds an immune system against misinformation could, if unchecked, serve as the ultimate surveillance tool. 
% Ultimately, the future of MAS lies not just in making agents that are smarter, but in making them \emph{intelligible}. The transition to Explicit Belief architectures is the critical step toward ensuring that the emerging multi-agent society empowers human agency through transparent, equitable, and verifiable exchange of beliefs.
\section{Conclusion}\label{sec:conclusion}

% We argue that a central obstacle to safe LLM-based Multi-Agent Systems is the lack of explicit, auditable belief representations, which prevents reliable distinction between benign assistance and strategic manipulation. To address this gap, we propose the \textbf{Explicit Belief Multi-Agent System (EB-MAS)} as a general-purpose cognitive substrate: by making belief states and updates explicit, EB-MAS enables explainability under imperfect information, provides a transparent trust interface for human--AI collaboration, and supports the analysis of collective belief dynamics at scale. Mechanisms such as \emph{Cognitive Firewalls} and \emph{Belief-Conditioned Action Synthesis} operationalize this vision by turning agents into verifiable epistemic actors, a prerequisite for safe and accountable multi-agent systems.

In this paper, we argue that safe LLM-powered multi-agent systems should treat \emph{auditable belief updates} as a core security boundary, rather than relying solely on output-level filtering or behavior-level policy optimization. EB-MAS operationalizes this view by externalizing belief states and update operations, providing explicit interfaces for evidence tracing, update bounding, and population-level propagation analysis. Through mechanisms such as \emph{Cognitive Firewalls} and \emph{Belief-Conditioned Action Synthesis}, EB-MAS makes belief-level influence more observable, constrainable, and accountable in open-world agent deployments.

% Authors are \textbf{required} to include a statement of the potential broader
% impact of their work, including its ethical aspects and future societal
% consequences. This statement should be in an unnumbered section at the end of
% the paper (co-located with Acknowledgements -- the two may appear in either
% order, but both must be before References), and does not count toward the paper
% page limit. In many cases, where the ethical impacts and expected societal
% implications are those that are well established when advancing the field of
% Machine Learning, substantial discussion is not required, and a simple
% statement such as the following will suffice:

% ``This paper presents work whose goal is to advance the field of Machine
% Learning. There are many potential societal consequences of our work, none
% which we feel must be specifically highlighted here.''

% The above statement can be used verbatim in such cases, but we encourage
% authors to think about whether there is content which does warrant further
% discussion, as this statement will be apparent if the paper is later flagged
% for ethics review.


\section*{Acknowledgment}
This work is partially funded by the European Union (under grant agreement ID 101212818). Views and opinions expressed are however those of the author(s) only and do not necessarily reflect those of the European Union or European Health and Digital Executive Agency (HADEA). Neither the European Union nor the granting authority can be held responsible for them. This work is partially supported by Innovate UK through AI-PASSPORT under Grant 10126404. This work was awarded a grant by the AI Security Institute (AISI) via the Alignment Project (Rare-Event Estimation in Large Language Models via Subset Simulation) and funded by EPSRC. Yi's contribution is partially supported through the Royal Society international exchanges programme and in part by the Engineering and Physical Sciences Research Council, through funding from RAi UK [EP/Y009800/1].

\bibliographystyle{plain}
\bibliography{example_paper}

@article{NING202473,
title = {A survey on multi-agent reinforcement learning and its application},
journal = {Journal of Automation and Intelligence},
volume = {3},
number = {2},
pages = {73-91},
year = {2024},
issn = {2949-8554},
doi = {https://doi.org/10.1016/j.jai.2024.02.003},
url = {https://www.sciencedirect.com/science/article/pii/S2949855424000042},
author = {Zepeng Ning and Lihua Xie}
}

@misc{fuchs2021theoryminddeepreinforcement,
      title={Theory of Mind for Deep Reinforcement Learning in Hanabi}, 
      author={Andrew Fuchs and Michael Walton and Theresa Chadwick and Doug Lange},
      year={2021},
      eprint={2101.09328},
      archivePrefix={arXiv},
      primaryClass={cs.AI},
      url={https://arxiv.org/abs/2101.09328}, 
}

@article{manenti2025beliefs,
	author = {Manenti, Giorgio L. and Khoneiveh, Sepideh and Gl{\"a}scher, Jan},
	title = {BELIEFS: A Hierarchical Theory of Mind Model based on Strategy Inference},
	elocation-id = {2025.06.25.660690},
	year = {2025},
	doi = {10.1101/2025.06.25.660690},
	publisher = {Cold Spring Harbor Laboratory},
	URL = {https://www.biorxiv.org/content/early/2025/12/02/2025.06.25.660690},
	eprint = {https://www.biorxiv.org/content/early/2025/12/02/2025.06.25.660690.full.pdf},
	journal = {bioRxiv}
}

@article{article,
author = {Shi, Haojun and Ye, Suyu and Fang, Xinyu and Jin, Chuanyang and Isik, Leyla and Kuo, Yen-Ling and Shu, Tianmin},
year = {2025},
month = {04},
pages = {1510-1519},
title = {MuMA-ToM: Multi-modal Multi-Agent Theory of Mind},
volume = {39},
journal = {Proceedings of the AAAI Conference on Artificial Intelligence},
doi = {10.1609/aaai.v39i2.32142}
}

@inproceedings{NEURIPS2024_861f7dad,
 author = {Motwani, Sumeet Ramesh and Baranchuk, Mikhail and Strohmeier, Martin and Bolina, Vijay and Torr, Philip H.S. and Hammond, Lewis and de Witt, Christian Schroeder},
 booktitle = {Advances in Neural Information Processing Systems},
 doi = {10.52202/079017-2336},
 editor = {A. Globerson and L. Mackey and D. Belgrave and A. Fan and U. Paquet and J. Tomczak and C. Zhang},
 pages = {73439--73486},
 publisher = {Curran Associates, Inc.},
 title = {Secret Collusion among AI Agents: Multi-Agent Deception via Steganography},
 url = {https://proceedings.neurips.cc/paper_files/paper/2024/file/861f7dad098aec1c3560fb7add468d41-Paper-Conference.pdf},
 volume = {37},
 year = {2024}
}

@inproceedings{f996313d83c74616938f1807b172eda1,
title = "Learning to Deceive in Multi-agent Hidden Role Games",
author = "Matthew Aitchison and Lyndon Benke and Penny Sweetser",
note = "Publisher Copyright: {\textcopyright} 2021, Springer Nature Switzerland AG.; 1st International Workshop on Deceptive AI, DeceptECAI 2020, held in conjunction with 24th European Conference on Artificial Intelligence, ECAI 2020 and 2nd International Workshop on Deceptive AI, DeceptAI 2021, held in conjunction with 30th International Joint Conference on Artificial Intelligence, IJCAI 2021 ; Conference date: 19-08-2020 Through 19-08-2020",
year = "2021",
doi = "10.1007/978-3-030-91779-1\_5",
language = "English",
isbn = "9783030917784",
series = "Communications in Computer and Information Science",
publisher = "Springer Science+Business Media B.V.",
pages = "55--75",
editor = "Stefan Sarkadi and Benjamin Wright and Peta Masters and Peter McBurney",
booktitle = "Deceptive AI - First International Workshop, DeceptECAI 2020 and Second International Workshop, DeceptAI 2021, Proceedings",
address = "Netherlands",
}

@misc{sun2022certifiablyrobustpolicylearning,
      title={Certifiably Robust Policy Learning against Adversarial Communication in Multi-agent Systems}, 
      author={Yanchao Sun and Ruijie Zheng and Parisa Hassanzadeh and Yongyuan Liang and Soheil Feizi and Sumitra Ganesh and Furong Huang},
      year={2022},
      eprint={2206.10158},
      archivePrefix={arXiv},
      primaryClass={cs.LG},
      url={https://arxiv.org/abs/2206.10158}, 
}

@inproceedings{10.5555/3535850.3536008,
author = {Xue, Wanqi and Qiu, Wei and An, Bo and Rabinovich, Zinovi and Obraztsova, Svetlana and Yeo, Chai Kiat},
title = {Mis-spoke or mis-lead: Achieving Robustness in Multi-Agent Communicative Reinforcement Learning},
year = {2022},
isbn = {9781450392136},
publisher = {International Foundation for Autonomous Agents and Multiagent Systems},
address = {Richland, SC},
booktitle = {Proceedings of the 21st International Conference on Autonomous Agents and Multiagent Systems},
pages = {1418–1426},
numpages = {9},
location = {Virtual Event, New Zealand},
series = {AAMAS '22}
}

@article{aimeur2023fake,
  title={Fake news, disinformation and misinformation in social media: a review},
  author={A{\"\i}meur, Esma and Amri, Sabrine and Brassard, Gilles},
  journal={Social Network Analysis and Mining},
  volume={13},
  number={1},
  pages={30},
  year={2023},
  publisher={Springer}
}

@inproceedings{10.1145/3442188.3445922,
author = {Bender, Emily M. and Gebru, Timnit and McMillan-Major, Angelina and Shmitchell, Shmargaret},
title = {On the Dangers of Stochastic Parrots: Can Language Models Be Too Big?},
year = {2021},
isbn = {9781450383097},
publisher = {Association for Computing Machinery},
address = {New York, NY, USA},
url = {https://doi.org/10.1145/3442188.3445922},
doi = {10.1145/3442188.3445922},
booktitle = {Proceedings of the 2021 ACM Conference on Fairness, Accountability, and Transparency},
pages = {610–623},
numpages = {14},
location = {Virtual Event, Canada},
series = {FAccT '21}
}

@book{goffman1974frame,
  title={Frame Analysis: An Essay on the Organization of Experience},
  author={Goffman, Erving},
  year={1974},
  publisher={Harvard University Press},
  address={Cambridge, MA}
}

@article{tversky1981framing,
  title={The Framing of Decisions and the Psychology of Choice},
  author={Tversky, Amos and Kahneman, Daniel},
  journal={Science},
  volume={211},
  number={4481},
  pages={453--458},
  year={1981},
  publisher={American Association for the Advancement of Science},
  doi={10.1126/science.7455683}
}

@article{Hagendorff_2024,
author = {Hagendorff, Thilo},
year = {2024},
month = {06},
pages = {e2317967121},
title = {Deception abilities emerged in large language models},
volume = {121},
journal = {Proceedings of the National Academy of Sciences of the United States of America},
doi = {10.1073/pnas.2317967121}
}

@inproceedings{park2023generative,
author = {Park, Joon Sung and O'Brien, Joseph and Cai, Carrie Jun and Morris, Meredith Ringel and Liang, Percy and Bernstein, Michael S.},
title = {Generative Agents: Interactive Simulacra of Human Behavior},
year = {2023},
isbn = {9798400701320},
publisher = {Association for Computing Machinery},
address = {New York, NY, USA},
url = {https://doi.org/10.1145/3586183.3606763},
doi = {10.1145/3586183.3606763},
booktitle = {Proceedings of the 36th Annual ACM Symposium on User Interface Software and Technology},
articleno = {2},
numpages = {22},
location = {San Francisco, CA, USA},
series = {UIST '23}
}

@article{argyle2023outofone, title={Out of One, Many: Using Language Models to Simulate Human Samples}, volume={31}, DOI={10.1017/pan.2023.2}, number={3}, journal={Political Analysis}, author={Argyle, Lisa P. and Busby, Ethan C. and Fulda, Nancy and Gubler, Joshua R. and Rytting, Christopher and Wingate, David}, year={2023}, pages={337–351}}

@inproceedings{horton2023economic,
author = {Filippas, Apostolos and Horton, John J. and Manning, Benjamin S.},
title = {Large Language Models as Simulated Economic Agents: What Can We Learn from Homo Silicus?},
year = {2024},
isbn = {9798400707049},
publisher = {Association for Computing Machinery},
address = {New York, NY, USA},
url = {https://doi.org/10.1145/3670865.3673513},
doi = {10.1145/3670865.3673513},
booktitle = {Proceedings of the 25th ACM Conference on Economics and Computation},
pages = {614–615},
numpages = {2},
location = {New Haven, CT, USA},
series = {EC '24}
}

@article{banisch2021opinion,
  title={Opinion dynamics and collective decisions},
  author={Lorenz, Jan and Neumann, Martin},
  journal={Advances in Complex Systems},
  volume={21},
  number={06n07},
  pages={1802002},
  year={2018},
  publisher={World Scientific}
}

@article{baumann2021echo,
  title={Modeling echo chambers and polarization dynamics in social networks},
  author={Baumann, Fabian and Lorenz-Spreen, Philipp and Sokolov, Igor M and Starnini, Michele},
  journal={Physical review letters},
  volume={124},
  number={4},
  pages={048301},
  year={2020},
  publisher={APS}
}

@inproceedings{
azmani2023cooperative,
title={Cooperative Foraging Behaviour Through Multi-Agent Reinforcement Learning with Graph-Based Communication},
author={Hicham Azmani and Andries Rosseau and Ann Nowe and Roxana R{\u{a}}dulescu},
booktitle={Sixteenth European Workshop on Reinforcement Learning},
year={2023},
url={https://openreview.net/forum?id=epzrEgNNSWL}
}

@inproceedings{li2023aiir,
  title={AIIR-MIX: Multi-agent reinforcement learning meets attention individual intrinsic reward mixing network},
  author={Li, Wei and Liu, Weiyan and Shao, Shitong and Huang, Shiyi},
  booktitle={Asian Conference on Machine Learning},
  pages={579--594},
  year={2023},
  organization={PMLR}
}

@inproceedings{10.5555/3545946.3598674,
author = {Zang, Yifan and He, Jinmin and Li, Kai and Fu, Haobo and Fu, Qiang and Xing, Junliang},
title = {Sequential Cooperative Multi-Agent Reinforcement Learning},
year = {2023},
isbn = {9781450394321},
publisher = {International Foundation for Autonomous Agents and Multiagent Systems},
address = {Richland, SC},
booktitle = {Proceedings of the 2023 International Conference on Autonomous Agents and Multiagent Systems},
pages = {485–493},
numpages = {9},
location = {London, United Kingdom},
series = {AAMAS '23}
}

@inproceedings{10.1609/aiide.v18i1.21954,
author = {Khan, Muhammad Junaid and Ahmed, Syed Hammad and Sukthankar, Gita},
title = {Transformer-based value function decomposition for cooperative multi-agent reinforcement learning in StarCraft},
year = {2022},
isbn = {978-1-57735-877-0},
publisher = {AAAI Press},
url = {https://doi.org/10.1609/aiide.v18i1.21954},
doi = {10.1609/aiide.v18i1.21954},
booktitle = {Proceedings of the Eighteenth AAAI Conference on Artificial Intelligence and Interactive Digital Entertainment},
articleno = {14},
numpages = {7},
location = {Pomona, CA, USA},
series = {AIIDE'22}
}

@inproceedings{10.5555/3709347.3743605,
author = {Geng, Minghong and Pateria, Shubham and Subagdja, Budhitama and Tan, Ah-Hwee},
title = {MOSMAC: A Multi-agent Reinforcement Learning Benchmark on Sequential Multi-Objective Tasks},
year = {2025},
isbn = {9798400714269},
publisher = {International Foundation for Autonomous Agents and Multiagent Systems},
address = {Richland, SC},
booktitle = {Proceedings of the 24th International Conference on Autonomous Agents and Multiagent Systems},
pages = {867–876},
numpages = {10},
location = {Detroit, MI, USA},
series = {AAMAS '25}
}

@misc{
xu2024language,
title={Language Agents with Reinforcement Learning for Strategic Play in the Werewolf Game},
author={Zelai Xu and Chao Yu and Fei Fang and Yu Wang and Yi Wu},
year={2024},
url={https://openreview.net/forum?id=N1gmpVd4iE}
}

@inproceedings{10.5555/3709347.3743819,
author = {Sarkar, Bidipta and Xia, Warren and Liu, C. Karen and Sadigh, Dorsa},
title = {Training Language Models for Social Deduction with Multi-Agent Reinforcement Learning},
year = {2025},
isbn = {9798400714269},
publisher = {International Foundation for Autonomous Agents and Multiagent Systems},
address = {Richland, SC},
booktitle = {Proceedings of the 24th International Conference on Autonomous Agents and Multiagent Systems},
pages = {1830–1839},
numpages = {10},
location = {Detroit, MI, USA},
series = {AAMAS '25}
}

@inproceedings{
xu2025learning,
title={Learning Strategic Language Agents in the Werewolf Game with Iterative Latent Space Policy Optimization},
author={Zelai Xu and Wanjun Gu and Chao Yu and Yi Wu and Yu Wang},
booktitle={Forty-second International Conference on Machine Learning},
year={2025},
url={https://openreview.net/forum?id=N2mOBiSqhc}
}

@inproceedings{10.1145/3544548.3581196,
author = {Jakesch, Maurice and Bhat, Advait and Buschek, Daniel and Zalmanson, Lior and Naaman, Mor},
title = {Co-Writing with Opinionated Language Models Affects Users’ Views},
year = {2023},
isbn = {9781450394215},
publisher = {Association for Computing Machinery},
address = {New York, NY, USA},
url = {https://doi.org/10.1145/3544548.3581196},
doi = {10.1145/3544548.3581196},
booktitle = {Proceedings of the 2023 CHI Conference on Human Factors in Computing Systems},
articleno = {111},
numpages = {15},
location = {Hamburg, Germany},
series = {CHI '23}
}

@inproceedings{10.1145/3706598.3713408,
author = {Danry, Valdemar and Pataranutaporn, Pat and Groh, Matthew and Epstein, Ziv},
title = {Deceptive Explanations by Large Language Models Lead People to Change their Beliefs About Misinformation More Often than Honest Explanations},
year = {2025},
isbn = {9798400713941},
publisher = {Association for Computing Machinery},
address = {New York, NY, USA},
url = {https://doi.org/10.1145/3706598.3713408},
doi = {10.1145/3706598.3713408},
booktitle = {Proceedings of the 2025 CHI Conference on Human Factors in Computing Systems},
articleno = {933},
numpages = {31},
location = {
},
series = {CHI '25}
}

@article{10.1145/3579592,
author = {Karinshak, Elise and Liu, Sunny Xun and Park, Joon Sung and Hancock, Jeffrey T.},
title = {Working With AI to Persuade: Examining a Large Language Model's Ability to Generate Pro-Vaccination Messages},
year = {2023},
issue_date = {April 2023},
publisher = {Association for Computing Machinery},
address = {New York, NY, USA},
volume = {7},
number = {CSCW1},
url = {https://doi.org/10.1145/3579592},
doi = {10.1145/3579592},
journal = {Proc. ACM Hum.-Comput. Interact.},
month = apr,
articleno = {116},
numpages = {29}
}

@inproceedings{10.1145/3613904.3642902,
author = {Tankelevitch, Lev and Kewenig, Viktor and Simkute, Auste and Scott, Ava Elizabeth and Sarkar, Advait and Sellen, Abigail and Rintel, Sean},
title = {The Metacognitive Demands and Opportunities of Generative AI},
year = {2024},
isbn = {9798400703300},
publisher = {Association for Computing Machinery},
address = {New York, NY, USA},
url = {https://doi.org/10.1145/3613904.3642902},
doi = {10.1145/3613904.3642902},
booktitle = {Proceedings of the 2024 CHI Conference on Human Factors in Computing Systems},
articleno = {680},
numpages = {24},
location = {Honolulu, HI, USA},
series = {CHI '24}
}

@article{osti_10463193,
place = {Country unknown/Code not available}, title = {Artificial intelligence in communication impacts language and social relationships}, url = {https://par.nsf.gov/biblio/10463193}, DOI = {10.1038/s41598-023-30938-9}, abstractNote = {Abstract Artificial intelligence (AI) is already widely used in daily communication, but despite concerns about AI’s negative effects on society the social consequences of using it to communicate remain largely unexplored. We investigate the social consequences of one of the most pervasive AI applications, algorithmic response suggestions (“smart replies”), which are used to send billions of messages each day. Two randomized experiments provide evidence that these types of algorithmic recommender systems change how people interact with and perceive one another in both pro-social and anti-social ways. We find that using algorithmic responses changes language and social relationships. More specifically, it increases communication speed, use of positive emotional language, and conversation partners evaluate each other as closer and more cooperative. However, consistent with common assumptions about the adverse effects of AI, people are evaluated more negatively if they are suspected to be using algorithmic responses. Thus, even though AI can increase the speed of communication and improve interpersonal perceptions, the prevailing anti-social connotations of AI undermine these potential benefits if used overtly.}, journal = {Scientific Reports}, volume = {13}, number = {1}, author = {Hohenstein, Jess and Kizilcec, Rene F. and DiFranzo, Dominic and Aghajari, Zhila and Mieczkowski, Hannah and Levy, Karen and Naaman, Mor and Hancock, Jeffrey and Jung, Malte F.}, year = {2023} }

@article{cheong2025epistemic,
  title={Epistemic and Emotional Harms of Generative AI: Towards Human-Centered First Amendment},
  author={Cheong, Inyoung},
  journal={Available at SSRN 5435335},
  year={2025}
}

@misc{kwesi2025exploringusersecurityprivacy,
      title={Exploring User Security and Privacy Attitudes and Concerns Toward the Use of General-Purpose LLM Chatbots for Mental Health}, 
      author={Jabari Kwesi and Jiaxun Cao and Riya Manchanda and Pardis Emami-Naeini},
      year={2025},
      eprint={2507.10695},
      archivePrefix={arXiv},
      primaryClass={cs.CY},
      url={https://arxiv.org/abs/2507.10695}, 
}

@inproceedings{chen-etal-2025-theory,
    title = "Theory of Mind in Large Language Models: Assessment and Enhancement",
    author = "Chen, Ruirui  and
      Jiang, Weifeng  and
      Qin, Chengwei  and
      Tan, Cheston",
    editor = "Che, Wanxiang  and
      Nabende, Joyce  and
      Shutova, Ekaterina  and
      Pilehvar, Mohammad Taher",
    booktitle = "Proceedings of the 63rd Annual Meeting of the Association for Computational Linguistics (Volume 1: Long Papers)",
    month = jul,
    year = "2025",
    address = "Vienna, Austria",
    publisher = "Association for Computational Linguistics",
    url = "https://aclanthology.org/2025.acl-long.1522/",
    doi = "10.18653/v1/2025.acl-long.1522",
    pages = "31539--31558",
    ISBN = "979-8-89176-251-0"
}

@article{ienca2023artificial,
  title={On artificial intelligence and manipulation},
  author={Ienca, Marcello},
  journal={Topoi},
  volume={42},
  number={3},
  pages={833--842},
  year={2023},
  publisher={Springer}
}

@inproceedings{srivastava-2025-large,
    title = "Large Language Models Threaten Language{'}s Epistemic and Communicative Foundations",
    author = "Srivastava, Shashank",
    editor = "Christodoulopoulos, Christos  and
      Chakraborty, Tanmoy  and
      Rose, Carolyn  and
      Peng, Violet",
    booktitle = "Proceedings of the 2025 Conference on Empirical Methods in Natural Language Processing",
    month = nov,
    year = "2025",
    address = "Suzhou, China",
    publisher = "Association for Computational Linguistics",
    url = "https://aclanthology.org/2025.emnlp-main.1457/",
    doi = "10.18653/v1/2025.emnlp-main.1457",
    pages = "28662--28676",
    ISBN = "979-8-89176-332-6"
}

@article{BEWA,
  title     = {{BEWA: A Bayesian Epistemology-Weighted Artificial Intelligence Framework for Scientific Inference}},
  author    = {Wright, Craig},
  journal   = {Journal of Engineering and Artificial Intelligence},
  volume    = {1},
  number    = {2},
  pages     = {1--32},
  year      = {2025},
  doi       = {10.64142/jeai.1.2.18},
  url       = {http://engineering-ai.academicsquare-pub.com/1/article/view/22}
}

@article{coeckelbergh2025ai,
  title={AI and epistemic agency: How AI influences belief revision and its normative implications},
  author={Coeckelbergh, Mark},
  journal={Social Epistemology},
  volume={40},
  number={1},
  pages={59--71},
  year={2026},
  publisher={Taylor \& Francis}
}

@article{mollema2025taxonomy,
  title={A taxonomy of epistemic injustice in the context of AI and the case for generative hermeneutical erasure},
  author={Mollema, Warmhold Jan Thomas},
  journal={AI and Ethics},
  volume={5},
  number={5},
  pages={5535--5555},
  year={2025},
  publisher={Springer}
}

@article{konigs2022artificial,
  title={Artificial intelligence and responsibility gaps: What is the problem?},
  author={K{\"o}nigs, Peter},
  journal={Ethics and Information Technology},
  volume={24},
  number={3},
  pages={36},
  year={2022},
  publisher={Springer}
}

@article{floridi2024ai,
author = {Floridi, Luciano},
year = {2025},
month = {02},
pages = {},
title = {AI as Agency without Intelligence: On Artificial Intelligence as a New Form of Artificial Agency and the Multiple Realisability of Agency Thesis},
volume = {38},
journal = {Philosophy \& Technology},
doi = {10.1007/s13347-025-00858-9}
}

@InProceedings{pmlr-v97-foerster19a,
  title = 	 {{B}ayesian Action Decoder for Deep Multi-Agent Reinforcement Learning},
  author =       {Foerster, Jakob and Song, Francis and Hughes, Edward and Burch, Neil and Dunning, Iain and Whiteson, Shimon and Botvinick, Matthew and Bowling, Michael},
  booktitle = 	 {Proceedings of the 36th International Conference on Machine Learning},
  pages = 	 {1942--1951},
  year = 	 {2019},
  editor = 	 {Chaudhuri, Kamalika and Salakhutdinov, Ruslan},
  volume = 	 {97},
  series = 	 {Proceedings of Machine Learning Research},
  month = 	 {09--15 Jun},
  publisher =    {PMLR},
  url = 	 {https://proceedings.mlr.press/v97/foerster19a.html}
}

@inproceedings{10.1609/aaai.v37i5.25755,
author = {Westby, Samuel and Riedl, Christoph},
title = {Collective intelligence in human-AI teams: a Bayesian theory of mind approach},
year = {2023},
isbn = {978-1-57735-880-0},
publisher = {AAAI Press},
url = {https://doi.org/10.1609/aaai.v37i5.25755},
doi = {10.1609/aaai.v37i5.25755},
booktitle = {Proceedings of the Thirty-Seventh AAAI Conference on Artificial Intelligence and Thirty-Fifth Conference on Innovative Applications of Artificial Intelligence and Thirteenth Symposium on Educational Advances in Artificial Intelligence},
articleno = {686},
numpages = {9},
series = {AAAI'23/IAAI'23/EAAI'23}
}

@inproceedings{NEURIPS2019_15212f24,
 author = {Khalvati, Koosha and Mirbagheri, Saghar and Park, Seongmin A. and Dreher, Jean-Claude and Rao, Rajesh PN},
 booktitle = {Advances in Neural Information Processing Systems},
 editor = {H. Wallach and H. Larochelle and A. Beygelzimer and F. d\textquotesingle Alch\'{e}-Buc and E. Fox and R. Garnett},
 pages = {},
 publisher = {Curran Associates, Inc.},
 title = {A Bayesian Theory of Conformity in Collective Decision Making},
 url = {https://proceedings.neurips.cc/paper_files/paper/2019/file/15212f24321aa2c3dc8e9acf820f3c15-Paper.pdf},
 volume = {32},
 year = {2019}
}

@inproceedings{10.5555/3709347.3743598,
author = {Friedrich, Paul and P\'{a}sztor, Barna and Ramponi, Giorgia},
title = {Learning Collusion in Episodic, Inventory-Constrained Markets},
year = {2025},
isbn = {9798400714269},
publisher = {International Foundation for Autonomous Agents and Multiagent Systems},
address = {Richland, SC},
booktitle = {Proceedings of the 24th International Conference on Autonomous Agents and Multiagent Systems},
pages = {803–812},
numpages = {10},
location = {Detroit, MI, USA},
series = {AAMAS '25}
}

@inproceedings{tam-etal-2023-evaluating, title = "Evaluating the Factual Consistency of Large Language Models Through News Summarization", author = "Tam, Derek and Mascarenhas, Anisha and Zhang, Shiyue and Kwan, Sarah and Bansal, Mohit and Raffel, Colin", editor = "Rogers, Anna and Boyd-Graber, Jordan and Okazaki, Naoaki", booktitle = "Findings of the Association for Computational Linguistics: ACL 2023", month = jul, year = "2023", address = "Toronto, Canada", publisher = "Association for Computational Linguistics", url = "https://aclanthology.org/2023.findings-acl.322/", doi = "10.18653/v1/2023.findings-acl.322", pages = "5220--5255" }

@article{cicero-2022,
  title={Human-level play in the game of Diplomacy by combining language models with strategic reasoning},
  author={FAIR and Bakhtin, Anton and Brown, Noam and Dinan, Emily and Farina, Gabriele and Flaherty, Colin and Fried, Daniel and Goff, Andrew and Gray, Jonathan and Hu, Hengyuan and others},
  journal={Science},
  volume={378},
  number={6624},
  pages={1067--1074},
  year={2022},
  publisher={American Association for the Advancement of Science}
}

@inproceedings{10.5555/3600270.3602737,
author = {Finkelstein, Mira and Liu, Lucy and Schlot, Nitsan Levy and Kolumbus, Yoav and Parkes, David C. and Rosenschein, Jeffrey S. and Keren, Sarah},
title = {Explainable reinforcement learning via model transforms},
year = {2022},
isbn = {9781713871088},
publisher = {Curran Associates Inc.},
address = {Red Hook, NY, USA},
booktitle = {Proceedings of the 36th International Conference on Neural Information Processing Systems},
articleno = {2467},
numpages = {13},
location = {New Orleans, LA, USA},
series = {NIPS '22}
}

@inproceedings{10.5555/3545946.3598627,
author = {Shi, Chongyang and Han, Shuo and Fu, Jie},
title = {Quantitative Planning with Action Deception in Concurrent Stochastic Games},
year = {2023},
isbn = {9781450394321},
publisher = {International Foundation for Autonomous Agents and Multiagent Systems},
address = {Richland, SC},
booktitle = {Proceedings of the 2023 International Conference on Autonomous Agents and Multiagent Systems},
pages = {122–130},
numpages = {9},
location = {London, United Kingdom},
series = {AAMAS '23}
}

@inproceedings{10.5555/3545946.3599097,
author = {Piazza, Nancirose and Behzadan, Vahid},
title = {A Theory of Mind Approach as Test-Time Mitigation Against Emergent Adversarial Communication},
year = {2023},
isbn = {9781450394321},
publisher = {International Foundation for Autonomous Agents and Multiagent Systems},
address = {Richland, SC},
booktitle = {Proceedings of the 2023 International Conference on Autonomous Agents and Multiagent Systems},
pages = {2842–2844},
numpages = {3},
location = {London, United Kingdom},
series = {AAMAS '23}
}

@inproceedings{10.5555/3545946.3599117,
author = {Ward, Francis Rhys and Toni, Francesca and Belardinelli, Francesco},
title = {Defining Deception in Structural Causal Games},
year = {2023},
isbn = {9781450394321},
publisher = {International Foundation for Autonomous Agents and Multiagent Systems},
address = {Richland, SC},
booktitle = {Proceedings of the 2023 International Conference on Autonomous Agents and Multiagent Systems},
pages = {2902–2904},
numpages = {3},
location = {London, United Kingdom},
series = {AAMAS '23}
}

@inproceedings{
schulz2023emergent,
title={Emergent deception and skepticism via theory of mind},
author={Lion Schulz and Nitay Alon and Jeffrey Rosenschein and Peter Dayan},
booktitle={First Workshop on Theory of Mind in Communicating Agents},
year={2023},
url={https://openreview.net/forum?id=yd8VOEpw8h}
}

@inproceedings{
orgad2025llms,
title={{LLM}s Know More Than They Show: On the Intrinsic Representation of {LLM} Hallucinations},
author={Hadas Orgad and Michael Toker and Zorik Gekhman and Roi Reichart and Idan Szpektor and Hadas Kotek and Yonatan Belinkov},
booktitle={The Thirteenth International Conference on Learning Representations},
year={2025},
url={https://openreview.net/forum?id=KRnsX5Em3W}
}

@misc{kadavath2022languagemodelsmostlyknow,
      title={Language Models (Mostly) Know What They Know}, 
      author={Saurav Kadavath and Tom Conerly and Amanda Askell and Tom Henighan and Dawn Drain and Ethan Perez and Nicholas Schiefer and Zac Hatfield-Dodds and Nova DasSarma and Eli Tran-Johnson and Scott Johnston and Sheer El-Showk and Andy Jones and Nelson Elhage and Tristan Hume and Anna Chen and Yuntao Bai and Sam Bowman and Stanislav Fort and Deep Ganguli and Danny Hernandez and Josh Jacobson and Jackson Kernion and Shauna Kravec and Liane Lovitt and Kamal Ndousse and Catherine Olsson and Sam Ringer and Dario Amodei and Tom Brown and Jack Clark and Nicholas Joseph and Ben Mann and Sam McCandlish and Chris Olah and Jared Kaplan},
      year={2022},
      eprint={2207.05221},
      archivePrefix={arXiv},
      primaryClass={cs.CL},
      url={https://arxiv.org/abs/2207.05221}, 
}

@inproceedings{long-etal-2025-truthful,
    title = "When Truthful Representations Flip Under Deceptive Instructions?",
    author = "Long, Xianxuan  and
      Fu, Yao  and
      Li, Runchao  and
      Sheng, Mu  and
      Yu, Haotian  and
      Han, Xiaotian  and
      Li, Pan",
    editor = "Christodoulopoulos, Christos  and
      Chakraborty, Tanmoy  and
      Rose, Carolyn  and
      Peng, Violet",
    booktitle = "Proceedings of the 2025 Conference on Empirical Methods in Natural Language Processing",
    month = nov,
    year = "2025",
    address = "Suzhou, China",
    publisher = "Association for Computational Linguistics",
    url = "https://aclanthology.org/2025.emnlp-main.826/",
    doi = "10.18653/v1/2025.emnlp-main.826",
    pages = "16315--16335",
    ISBN = "979-8-89176-332-6"
}

@inproceedings{dai-etal-2022-knowledge,
    title = "Knowledge Neurons in Pretrained Transformers",
    author = "Dai, Damai  and
      Dong, Li  and
      Hao, Yaru  and
      Sui, Zhifang  and
      Chang, Baobao  and
      Wei, Furu",
    editor = "Muresan, Smaranda  and
      Nakov, Preslav  and
      Villavicencio, Aline",
    booktitle = "Proceedings of the 60th Annual Meeting of the Association for Computational Linguistics (Volume 1: Long Papers)",
    month = may,
    year = "2022",
    address = "Dublin, Ireland",
    publisher = "Association for Computational Linguistics",
    url = "https://aclanthology.org/2022.acl-long.581/",
    doi = "10.18653/v1/2022.acl-long.581",
    pages = "8493--8502"
}

@misc{elhage2022toymodelssuperposition,
      title={Toy Models of Superposition}, 
      author={Nelson Elhage and Tristan Hume and Catherine Olsson and Nicholas Schiefer and Tom Henighan and Shauna Kravec and Zac Hatfield-Dodds and Robert Lasenby and Dawn Drain and Carol Chen and Roger Grosse and Sam McCandlish and Jared Kaplan and Dario Amodei and Martin Wattenberg and Christopher Olah},
      year={2022},
      eprint={2209.10652},
      archivePrefix={arXiv},
      primaryClass={cs.LG},
      url={https://arxiv.org/abs/2209.10652}, 
}

@article{das2025tracealign,
  title={TRACEALIGN--Tracing the Drift: Attributing Alignment Failures to Training-Time Belief Sources in LLMs},
  author={Das, Amitava and Jain, Vinija and Chadha, Aman},
  journal={arXiv preprint arXiv:2508.02063},
  year={2025}
}

@article{cohen-etal-2024-evaluating,
    title = "Evaluating the Ripple Effects of Knowledge Editing in Language Models",
    author = "Cohen, Roi  and
      Biran, Eden  and
      Yoran, Ori  and
      Globerson, Amir  and
      Geva, Mor",
    journal = "Transactions of the Association for Computational Linguistics",
    volume = "12",
    year = "2024",
    address = "Cambridge, MA",
    publisher = "MIT Press",
    url = "https://aclanthology.org/2024.tacl-1.16/",
    doi = "10.1162/tacl_a_00644",
    pages = "283--298"
}

@inproceedings{
sharma2024towards,
title={Towards Understanding Sycophancy in Language Models},
author={Mrinank Sharma and Meg Tong and Tomasz Korbak and David Duvenaud and Amanda Askell and Samuel R. Bowman and Esin DURMUS and Zac Hatfield-Dodds and Scott R Johnston and Shauna M Kravec and Timothy Maxwell and Sam McCandlish and Kamal Ndousse and Oliver Rausch and Nicholas Schiefer and Da Yan and Miranda Zhang and Ethan Perez},
booktitle={The Twelfth International Conference on Learning Representations},
year={2024},
url={https://openreview.net/forum?id=tvhaxkMKAn}
}

@inproceedings{10.5555/3766078.3766275,
author = {Zou, Wei and Geng, Runpeng and Wang, Binghui and Jia, Jinyuan},
title = {PoisonedRAG: knowledge corruption attacks to retrieval-augmented generation of large language models},
year = {2025},
isbn = {978-1-939133-52-6},
publisher = {USENIX Association},
address = {USA},
booktitle = {Proceedings of the 34th USENIX Conference on Security Symposium},
articleno = {197},
numpages = {18},
location = {Seattle, WA, USA},
series = {SEC '25}
}

@inproceedings{
kirk2024understanding,
title={Understanding the Effects of {RLHF} on {LLM} Generalisation and Diversity},
author={Robert Kirk and Ishita Mediratta and Christoforos Nalmpantis and Jelena Luketina and Eric Hambro and Edward Grefenstette and Roberta Raileanu},
booktitle={The Twelfth International Conference on Learning Representations},
year={2024},
url={https://openreview.net/forum?id=PXD3FAVHJT}
}

@inproceedings{
coste2024reward,
title={Reward Model Ensembles Help Mitigate Overoptimization},
author={Thomas Coste and Usman Anwar and Robert Kirk and David Krueger},
booktitle={The Twelfth International Conference on Learning Representations},
year={2024},
url={https://openreview.net/forum?id=dcjtMYkpXx}
}

@inproceedings{zhan-etal-2024-injecagent,
    title = "{I}njec{A}gent: Benchmarking Indirect Prompt Injections in Tool-Integrated Large Language Model Agents",
    author = "Zhan, Qiusi  and
      Liang, Zhixiang  and
      Ying, Zifan  and
      Kang, Daniel",
    editor = "Ku, Lun-Wei  and
      Martins, Andre  and
      Srikumar, Vivek",
    booktitle = "Findings of the Association for Computational Linguistics: ACL 2024",
    month = aug,
    year = "2024",
    address = "Bangkok, Thailand",
    publisher = "Association for Computational Linguistics",
    url = "https://aclanthology.org/2024.findings-acl.624/",
    doi = "10.18653/v1/2024.findings-acl.624",
    pages = "10471--10506"
}

@inproceedings{
zhang2025agent,
title={Agent Security Bench ({ASB}): Formalizing and Benchmarking Attacks and Defenses in {LLM}-based Agents},
author={Hanrong Zhang and Jingyuan Huang and Kai Mei and Yifei Yao and Zhenting Wang and Chenlu Zhan and Hongwei Wang and Yongfeng Zhang},
booktitle={The Thirteenth International Conference on Learning Representations},
year={2025},
url={https://openreview.net/forum?id=V4y0CpX4hK}
}

@inproceedings{zhan-etal-2025-adaptive,
    title = "Adaptive Attacks Break Defenses Against Indirect Prompt Injection Attacks on {LLM} Agents",
    author = "Zhan, Qiusi  and
      Fang, Richard  and
      Panchal, Henil Shalin  and
      Kang, Daniel",
    editor = "Chiruzzo, Luis  and
      Ritter, Alan  and
      Wang, Lu",
    booktitle = "Findings of the Association for Computational Linguistics: NAACL 2025",
    month = apr,
    year = "2025",
    address = "Albuquerque, New Mexico",
    publisher = "Association for Computational Linguistics",
    url = "https://aclanthology.org/2025.findings-naacl.395/",
    doi = "10.18653/v1/2025.findings-naacl.395",
    pages = "7101--7117",
    ISBN = "979-8-89176-195-7"
}

@inproceedings{
yang2024watch,
title={Watch Out for Your Agents! Investigating Backdoor Threats to {LLM}-Based Agents},
author={Wenkai Yang and Xiaohan Bi and Yankai Lin and Sishuo Chen and Jie Zhou and Xu Sun},
booktitle={The Thirty-eighth Annual Conference on Neural Information Processing Systems},
year={2024},
url={https://openreview.net/forum?id=Nf4MHF1pi5}
}

@inproceedings{
motwani2024secret,
title={Secret Collusion among {AI} Agents: Multi-Agent Deception via Steganography},
author={Sumeet Ramesh Motwani and Mikhail Baranchuk and Martin Strohmeier and Vijay Bolina and Philip Torr and Lewis Hammond and Christian Schroeder de Witt},
booktitle={The Thirty-eighth Annual Conference on Neural Information Processing Systems},
year={2024},
url={https://openreview.net/forum?id=bnNSQhZJ88}
}

@article{chen2025aiagentbehavioralscience,
  title={AI agent behavioral science},
  author={Chen, Lin and Zhang, Yunke and Feng, Jie and Chai, Haoye and Zhang, Honglin and Fan, Bingbing and Ma, Yibo and Zhang, Shiyuan and Li, Nian and Liu, Tianhui and others},
  journal={Humanities and Social Sciences Communications},
  year={2026},
  publisher={Palgrave}
}

@misc{xue2024badragidentifyingvulnerabilitiesretrieval,
      title={BadRAG: Identifying Vulnerabilities in Retrieval Augmented Generation of Large Language Models}, 
      author={Jiaqi Xue and Mengxin Zheng and Yebowen Hu and Fei Liu and Xun Chen and Qian Lou},
      year={2024},
      eprint={2406.00083},
      archivePrefix={arXiv},
      primaryClass={cs.CR},
      url={https://arxiv.org/abs/2406.00083}, 
}

@inproceedings{10.5555/3766078.3766273,
author = {Shafran, Avital and Schuster, Roei and Shmatikov, Vitaly},
title = {Machine against the RAG: jamming retrieval-augmented generation with blocker documents},
year = {2025},
isbn = {978-1-939133-52-6},
publisher = {USENIX Association},
address = {USA},
booktitle = {Proceedings of the 34th USENIX Conference on Security Symposium},
articleno = {195},
numpages = {20},
location = {Seattle, WA, USA},
series = {SEC '25}
}

@inproceedings{tan-etal-2025-revprag,
    title = "{R}ev{PRAG}: Revealing Poisoning Attacks in Retrieval-Augmented Generation through {LLM} Activation Analysis",
    author = "Tan, Xue  and
      Luan, Hao  and
      Luo, Mingyu  and
      Sun, Xiaoyan  and
      Chen, Ping  and
      Dai, Jun",
    editor = "Christodoulopoulos, Christos  and
      Chakraborty, Tanmoy  and
      Rose, Carolyn  and
      Peng, Violet",
    booktitle = "Findings of the Association for Computational Linguistics: EMNLP 2025",
    month = nov,
    year = "2025",
    address = "Suzhou, China",
    publisher = "Association for Computational Linguistics",
    url = "https://aclanthology.org/2025.findings-emnlp.698/",
    doi = "10.18653/v1/2025.findings-emnlp.698",
    pages = "12999--13011",
    ISBN = "979-8-89176-335-7"
}

@misc{debenedetti2024agentdojodynamicenvironmentevaluate,
      title={AgentDojo: A Dynamic Environment to Evaluate Prompt Injection Attacks and Defenses for LLM Agents}, 
      author={Edoardo Debenedetti and Jie Zhang and Mislav Balunović and Luca Beurer-Kellner and Marc Fischer and Florian Tramèr},
      year={2024},
      eprint={2406.13352},
      archivePrefix={arXiv},
      primaryClass={cs.CR},
      url={https://arxiv.org/abs/2406.13352}, 
}

@inproceedings{NEURIPS2024_63092d79,
 author = {Chao, Patrick and Debenedetti, Edoardo and Robey, Alexander and Andriushchenko, Maksym and Croce, Francesco and Sehwag, Vikash and Dobriban, Edgar and Flammarion, Nicolas and Pappas, George J. and Tram\`{e}r, Florian and Hassani, Hamed and Wong, Eric},
 booktitle = {Advances in Neural Information Processing Systems},
 doi = {10.52202/079017-1745},
 editor = {A. Globerson and L. Mackey and D. Belgrave and A. Fan and U. Paquet and J. Tomczak and C. Zhang},
 pages = {55005--55029},
 publisher = {Curran Associates, Inc.},
 title = {JailbreakBench: An Open Robustness Benchmark for Jailbreaking Large Language Models},
 url = {https://proceedings.neurips.cc/paper_files/paper/2024/file/63092d79154adebd7305dfd498cbff70-Paper-Datasets_and_Benchmarks_Track.pdf},
 volume = {37},
 year = {2024}
}

@misc{ge2025backdoorsspeakunderstandingllm,
      title={When Backdoors Speak: Understanding LLM Backdoor Attacks Through Model-Generated Explanations}, 
      author={Huaizhi Ge and Yiming Li and Qifan Wang and Yongfeng Zhang and Ruixiang Tang},
      year={2025},
      eprint={2411.12701},
      archivePrefix={arXiv},
      primaryClass={cs.CR},
      url={https://arxiv.org/abs/2411.12701}, 
}

@misc{hubinger2024sleeperagentstrainingdeceptive,
      title={Sleeper Agents: Training Deceptive LLMs that Persist Through Safety Training}, 
      author={Evan Hubinger and Carson Denison and Jesse Mu and Mike Lambert and Meg Tong and Monte MacDiarmid and Tamera Lanham and Daniel M. Ziegler and Tim Maxwell and Newton Cheng and Adam Jermyn and Amanda Askell and Ansh Radhakrishnan and Cem Anil and David Duvenaud and Deep Ganguli and Fazl Barez and Jack Clark and Kamal Ndousse and Kshitij Sachan and Michael Sellitto and Mrinank Sharma and Nova DasSarma and Roger Grosse and Shauna Kravec and Yuntao Bai and Zachary Witten and Marina Favaro and Jan Brauner and Holden Karnofsky and Paul Christiano and Samuel R. Bowman and Logan Graham and Jared Kaplan and Sören Mindermann and Ryan Greenblatt and Buck Shlegeris and Nicholas Schiefer and Ethan Perez},
      year={2024},
      eprint={2401.05566},
      archivePrefix={arXiv},
      primaryClass={cs.CR},
      url={https://arxiv.org/abs/2401.05566}, 
}

@inproceedings{wu-etal-2024-shall,
    title = "Shall We Team Up: Exploring Spontaneous Cooperation of Competing {LLM} Agents",
    author = "Wu, Zengqing  and
      Peng, Run  and
      Zheng, Shuyuan  and
      Liu, Qianying  and
      Han, Xu  and
      Kwon, Brian I.  and
      Onizuka, Makoto  and
      Tang, Shaojie  and
      Xiao, Chuan",
    editor = "Al-Onaizan, Yaser  and
      Bansal, Mohit  and
      Chen, Yun-Nung",
    booktitle = "Findings of the Association for Computational Linguistics: EMNLP 2024",
    month = nov,
    year = "2024",
    address = "Miami, Florida, USA",
    publisher = "Association for Computational Linguistics",
    url = "https://aclanthology.org/2024.findings-emnlp.297/",
    doi = "10.18653/v1/2024.findings-emnlp.297",
    pages = "5163--5186"
}

@misc{fish2025algorithmiccollusionlargelanguage,
      title={Algorithmic Collusion by Large Language Models}, 
      author={Sara Fish and Yannai A. Gonczarowski and Ran I. Shorrer},
      year={2026},
      eprint={2404.00806},
      archivePrefix={arXiv},
      primaryClass={econ.GN},
      url={https://arxiv.org/abs/2404.00806}, 
}

@misc{lin2025strategiccollusionllmagents,
      title={Strategic Collusion of LLM Agents: Market Division in Multi-Commodity Competitions}, 
      author={Ryan Y. Lin and Siddhartha Ojha and Kevin Cai and Maxwell F. Chen},
      year={2025},
      eprint={2410.00031},
      archivePrefix={arXiv},
      primaryClass={cs.GT},
      url={https://arxiv.org/abs/2410.00031}, 
}

@misc{tailor2025auditwhisperdetectingsteganographic,
      title={Audit the Whisper: Detecting Steganographic Collusion in Multi-Agent LLMs}, 
      author={Om Tailor},
      year={2025},
      eprint={2510.04303},
      archivePrefix={arXiv},
      primaryClass={cs.MA},
      url={https://arxiv.org/abs/2510.04303}, 
}

@inproceedings{
ghaemi2025a,
title={A Survey of Collusion Risk in {LLM}-Powered Multi-Agent Systems},
author={Mohammad Sajjad Ghaemi},
booktitle={Socially Responsible and Trustworthy Foundation Models at NeurIPS 2025},
year={2025},
url={https://openreview.net/forum?id=Ylh8617Qyd}
}

@inproceedings{wu-etal-2024-enhancing-training,
    title = "Enhancing Training Data Attribution for Large Language Models with Fitting Error Consideration",
    author = "Wu, Kangxi  and
      Pang, Liang  and
      Shen, Huawei  and
      Cheng, Xueqi",
    editor = "Al-Onaizan, Yaser  and
      Bansal, Mohit  and
      Chen, Yun-Nung",
    booktitle = "Proceedings of the 2024 Conference on Empirical Methods in Natural Language Processing",
    month = nov,
    year = "2024",
    address = "Miami, Florida, USA",
    publisher = "Association for Computational Linguistics",
    url = "https://aclanthology.org/2024.emnlp-main.782/",
    doi = "10.18653/v1/2024.emnlp-main.782",
    pages = "14131--14143"
}

@inproceedings{qin-etal-2024-new,
    title = "Why Does New Knowledge Create Messy Ripple Effects in {LLM}s?",
    author = "Qin, Jiaxin  and
      Zhang, Zixuan  and
      Han, Chi  and
      Yu, Pengfei  and
      Li, Manling  and
      Ji, Heng",
    editor = "Al-Onaizan, Yaser  and
      Bansal, Mohit  and
      Chen, Yun-Nung",
    booktitle = "Proceedings of the 2024 Conference on Empirical Methods in Natural Language Processing",
    month = nov,
    year = "2024",
    address = "Miami, Florida, USA",
    publisher = "Association for Computational Linguistics",
    url = "https://aclanthology.org/2024.emnlp-main.700/",
    doi = "10.18653/v1/2024.emnlp-main.700",
    pages = "12602--12609"
}

@inproceedings{sankararaman-etal-2024-provenance,
    title = "Provenance: A Light-weight Fact-checker for Retrieval Augmented {LLM} Generation Output",
    author = "Sankararaman, Hithesh  and
      Yasin, Mohammed Nasheed  and
      Sorensen, Tanner  and
      Bari, Alessandro Di  and
      Stolcke, Andreas",
    editor = "Dernoncourt, Franck  and
      Preo{\c{t}}iuc-Pietro, Daniel  and
      Shimorina, Anastasia",
    booktitle = "Proceedings of the 2024 Conference on Empirical Methods in Natural Language Processing: Industry Track",
    month = nov,
    year = "2024",
    address = "Miami, Florida, US",
    publisher = "Association for Computational Linguistics",
    url = "https://aclanthology.org/2024.emnlp-industry.97/",
    doi = "10.18653/v1/2024.emnlp-industry.97",
    pages = "1305--1313"
}

@inproceedings{
huben2024sparse,
title={Sparse Autoencoders Find Highly Interpretable Features in Language Models},
author={Robert Huben and Hoagy Cunningham and Logan Riggs Smith and Aidan Ewart and Lee Sharkey},
booktitle={The Twelfth International Conference on Learning Representations},
year={2024},
url={https://openreview.net/forum?id=F76bwRSLeK}
}

@inproceedings{su-etal-2024-unsupervised,
    title = "Unsupervised Real-Time Hallucination Detection based on the Internal States of Large Language Models",
    author = "Su, Weihang  and
      Wang, Changyue  and
      Ai, Qingyao  and
      Hu, Yiran  and
      Wu, Zhijing  and
      Zhou, Yujia  and
      Liu, Yiqun",
    editor = "Ku, Lun-Wei  and
      Martins, Andre  and
      Srikumar, Vivek",
    booktitle = "Findings of the Association for Computational Linguistics: ACL 2024",
    month = aug,
    year = "2024",
    address = "Bangkok, Thailand",
    publisher = "Association for Computational Linguistics",
    url = "https://aclanthology.org/2024.findings-acl.854/",
    doi = "10.18653/v1/2024.findings-acl.854",
    pages = "14379--14391"
}

@inproceedings{ji-etal-2024-llm,
    title = "{LLM} Internal States Reveal Hallucination Risk Faced With a Query",
    author = "Ji, Ziwei  and
      Chen, Delong  and
      Ishii, Etsuko  and
      Cahyawijaya, Samuel  and
      Bang, Yejin  and
      Wilie, Bryan  and
      Fung, Pascale",
    editor = "Belinkov, Yonatan  and
      Kim, Najoung  and
      Jumelet, Jaap  and
      Mohebbi, Hosein  and
      Mueller, Aaron  and
      Chen, Hanjie",
    booktitle = "Proceedings of the 7th BlackboxNLP Workshop: Analyzing and Interpreting Neural Networks for NLP",
    month = nov,
    year = "2024",
    address = "Miami, Florida, US",
    publisher = "Association for Computational Linguistics",
    url = "https://aclanthology.org/2024.blackboxnlp-1.6/",
    doi = "10.18653/v1/2024.blackboxnlp-1.6",
    pages = "88--104"
}

@inproceedings{
dunefsky2024transcoders,
title={Transcoders find interpretable {LLM} feature circuits},
author={Jacob Dunefsky and Philippe Chlenski and Neel Nanda},
booktitle={The Thirty-eighth Annual Conference on Neural Information Processing Systems},
year={2024},
url={https://openreview.net/forum?id=J6zHcScAo0}
}

@inproceedings{zhao-etal-2025-steering,
    title = "Steering Knowledge Selection Behaviours in {LLM}s via {SAE}-Based Representation Engineering",
    author = "Zhao, Yu  and
      Devoto, Alessio  and
      Hong, Giwon  and
      Du, Xiaotang  and
      Gema, Aryo Pradipta  and
      Wang, Hongru  and
      He, Xuanli  and
      Wong, Kam-Fai  and
      Minervini, Pasquale",
    editor = "Chiruzzo, Luis  and
      Ritter, Alan  and
      Wang, Lu",
    booktitle = "Proceedings of the 2025 Conference of the Nations of the Americas Chapter of the Association for Computational Linguistics: Human Language Technologies (Volume 1: Long Papers)",
    month = apr,
    year = "2025",
    address = "Albuquerque, New Mexico",
    publisher = "Association for Computational Linguistics",
    url = "https://aclanthology.org/2025.naacl-long.264/",
    doi = "10.18653/v1/2025.naacl-long.264",
    pages = "5117--5136",
    ISBN = "979-8-89176-189-6"
}

@inproceedings{10.5555/3737916.3740051,
author = {Delfosse, Quentin and Sztwiertnia, Sebastian and Rothermel, Mark and Stammer, Wolfgang and Kersting, Kristian},
title = {Interpretable concept bottlenecks to align reinforcement learning agents},
year = {2024},
isbn = {9798331314385},
publisher = {Curran Associates Inc.},
address = {Red Hook, NY, USA},
booktitle = {Proceedings of the 38th International Conference on Neural Information Processing Systems},
articleno = {2135},
numpages = {30},
location = {Vancouver, BC, Canada},
series = {NIPS '24}
}

@inproceedings{10.1609/aaai.v38i13.29372,
author = {Patel, Sagar and Jyothi, Sangeetha Abdu and Narodytska, Nina},
title = {CrystalBox: future-based explanations for input-driven deep RL systems},
year = {2024},
isbn = {978-1-57735-887-9},
publisher = {AAAI Press},
url = {https://doi.org/10.1609/aaai.v38i13.29372},
doi = {10.1609/aaai.v38i13.29372},
booktitle = {Proceedings of the Thirty-Eighth AAAI Conference on Artificial Intelligence and Thirty-Sixth Conference on Innovative Applications of Artificial Intelligence and Fourteenth Symposium on Educational Advances in Artificial Intelligence},
articleno = {1624},
numpages = {9},
series = {AAAI'24/IAAI'24/EAAI'24}
}

@inproceedings{NEURIPS2024_a995960d,
 author = {Lang, Leon and Foote, Davis and Russell, Stuart and Dragan, Anca and Jenner, Erik and Emmons, Scott},
 booktitle = {Advances in Neural Information Processing Systems},
 doi = {10.52202/079017-2959},
 editor = {A. Globerson and L. Mackey and D. Belgrave and A. Fan and U. Paquet and J. Tomczak and C. Zhang},
 pages = {93240--93299},
 publisher = {Curran Associates, Inc.},
 title = {When Your AIs Deceive You: Challenges of Partial Observability in Reinforcement Learning from Human Feedback},
 url = {https://proceedings.neurips.cc/paper_files/paper/2024/file/a995960dd0193654d6b18eca4ac5b936-Paper-Conference.pdf},
 volume = {37},
 year = {2024}
}

@inproceedings{NEURIPS2024_eb113910,
 author = {Chen, Zhaorun and Xiang, Zhen and Xiao, Chaowei and Song, Dawn and Li, Bo},
 booktitle = {Advances in Neural Information Processing Systems},
 doi = {10.52202/079017-4136},
 editor = {A. Globerson and L. Mackey and D. Belgrave and A. Fan and U. Paquet and J. Tomczak and C. Zhang},
 pages = {130185--130213},
 publisher = {Curran Associates, Inc.},
 title = {AgentPoison: Red-teaming LLM Agents via Poisoning Memory or Knowledge Bases},
 url = {https://proceedings.neurips.cc/paper_files/paper/2024/file/eb113910e9c3f6242541c1652e30dfd6-Paper-Conference.pdf},
 volume = {37},
 year = {2024}
}

@inproceedings{ijcai2025p1175,
  title     = {A Survey on the Feedback Mechanism of LLM-based AI Agents},
  author    = {Liu, Zhipeng and Bai, Xuefeng and Chen, Kehai and Chen, Xinyang and Li, Xiucheng and Xiang, Yang and Liu, Jin and Li, Hong-Dong and Wang, Yaowei and Nie, Liqiang and Zhang, Min},
  booktitle = {Proceedings of the Thirty-Fourth International Joint Conference on
               Artificial Intelligence, {IJCAI-25}},
  publisher = {International Joint Conferences on Artificial Intelligence Organization},
  editor    = {James Kwok},
  pages     = {10582--10592},
  year      = {2025},
  month     = {8},
  note      = {Survey Track},
  doi       = {10.24963/ijcai.2025/1175},
  url       = {https://doi.org/10.24963/ijcai.2025/1175},
}

@inproceedings{chen-etal-2024-reconcile,
    title = "{R}e{C}oncile: Round-Table Conference Improves Reasoning via Consensus among Diverse {LLM}s",
    author = "Chen, Justin  and
      Saha, Swarnadeep  and
      Bansal, Mohit",
    editor = "Ku, Lun-Wei  and
      Martins, Andre  and
      Srikumar, Vivek",
    booktitle = "Proceedings of the 62nd Annual Meeting of the Association for Computational Linguistics (Volume 1: Long Papers)",
    month = aug,
    year = "2024",
    address = "Bangkok, Thailand",
    publisher = "Association for Computational Linguistics",
    url = "https://aclanthology.org/2024.acl-long.381/",
    doi = "10.18653/v1/2024.acl-long.381",
    pages = "7066--7085"
}

@article{bikhchandani1992theory,
  title={A theory of fads, fashion, custom, and cultural change as informational cascades},
  author={Bikhchandani, Sushil and Hirshleifer, David and Welch, Ivo},
  journal={Journal of political Economy},
  volume={100},
  number={5},
  pages={992--1026},
  year={1992},
  publisher={The University of Chicago Press}
}

@article{centola2010spread,
  title={The spread of behavior in an online social network experiment},
  author={Centola, Damon},
  journal={science},
  volume={329},
  number={5996},
  pages={1194--1197},
  year={2010},
  publisher={American Association for the Advancement of Science}
}

@article{vosoughi2018spread,
  title={The spread of true and false news online},
  author={Vosoughi, Soroush and Roy, Deb and Aral, Sinan},
  journal={science},
  volume={359},
  number={6380},
  pages={1146--1151},
  year={2018},
  publisher={American Association for the Advancement of Science}
}

@article{
lazer2018science,
author = {David M. J. Lazer  and Matthew A. Baum  and Yochai Benkler  and Adam J. Berinsky  and Kelly M. Greenhill  and Filippo Menczer  and Miriam J. Metzger  and Brendan Nyhan  and Gordon Pennycook  and David Rothschild  and Michael Schudson  and Steven A. Sloman  and Cass R. Sunstein  and Emily A. Thorson  and Duncan J. Watts  and Jonathan L. Zittrain },
title = {The science of fake news},
journal = {Science},
volume = {359},
number = {6380},
pages = {1094-1096},
year = {2018},
doi = {10.1126/science.aao2998},
URL = {https://www.science.org/doi/abs/10.1126/science.aao2998},
eprint = {https://www.science.org/doi/pdf/10.1126/science.aao2998}}

@inproceedings{shen-etal-2025-understanding,
    title = "Understanding the Information Propagation Effects of Communication Topologies in {LLM}-based Multi-Agent Systems",
    author = "Shen, Xu  and
      Liu, Yixin  and
      Dai, Yiwei  and
      Wang, Yili  and
      Miao, Rui  and
      Tan, Yue  and
      Pan, Shirui  and
      Wang, Xin",
    editor = "Christodoulopoulos, Christos  and
      Chakraborty, Tanmoy  and
      Rose, Carolyn  and
      Peng, Violet",
    booktitle = "Proceedings of the 2025 Conference on Empirical Methods in Natural Language Processing",
    month = nov,
    year = "2025",
    address = "Suzhou, China",
    publisher = "Association for Computational Linguistics",
    url = "https://aclanthology.org/2025.emnlp-main.623/",
    doi = "10.18653/v1/2025.emnlp-main.623",
    pages = "12347--12361",
    ISBN = "979-8-89176-332-6"
}

@inproceedings{hu-etal-2025-removal,
    title = "Removal of Hallucination on Hallucination: Debate-Augmented {RAG}",
    author = "Hu, Wentao  and
      Zhang, Wengyu  and
      Jiang, Yiyang  and
      Zhang, Chen Jason  and
      Wei, Xiaoyong  and
      Qing, Li",
    editor = "Che, Wanxiang  and
      Nabende, Joyce  and
      Shutova, Ekaterina  and
      Pilehvar, Mohammad Taher",
    booktitle = "Proceedings of the 63rd Annual Meeting of the Association for Computational Linguistics (Volume 1: Long Papers)",
    month = jul,
    year = "2025",
    address = "Vienna, Austria",
    publisher = "Association for Computational Linguistics",
    url = "https://aclanthology.org/2025.acl-long.770/",
    doi = "10.18653/v1/2025.acl-long.770",
    pages = "15839--15853",
    ISBN = "979-8-89176-251-0"
}

@misc{zhang2026practicalpoisoningattacksretrievalaugmented,
      title={Practical Poisoning Attacks against Retrieval-Augmented Generation}, 
      author={Baolei Zhang and Yuxi Chen and Zhuqing Liu and Lihai Nie and Tong Li and Zheli Liu and Minghong Fang},
      year={2026},
      eprint={2504.03957},
      archivePrefix={arXiv},
      primaryClass={cs.CR},
      url={https://arxiv.org/abs/2504.03957}, 
}

@inproceedings{chang-etal-2025-one,
    title = "One Shot Dominance: Knowledge Poisoning Attack on Retrieval-Augmented Generation Systems",
    author = "Chang, Zhiyuan  and
      Li, Mingyang  and
      Jia, Xiaojun  and
      Wang, Junjie  and
      Huang, Yuekai  and
      Jiang, Ziyou  and
      Liu, Yang  and
      Wang, Qing",
    editor = "Christodoulopoulos, Christos  and
      Chakraborty, Tanmoy  and
      Rose, Carolyn  and
      Peng, Violet",
    booktitle = "Findings of the Association for Computational Linguistics: EMNLP 2025",
    month = nov,
    year = "2025",
    address = "Suzhou, China",
    publisher = "Association for Computational Linguistics",
    url = "https://aclanthology.org/2025.findings-emnlp.1023/",
    doi = "10.18653/v1/2025.findings-emnlp.1023",
    pages = "18811--18825",
    ISBN = "979-8-89176-335-7"
}

@inproceedings{
zhang2025poisonedeye,
title={PoisonedEye: Knowledge Poisoning Attack on Retrieval-Augmented Generation based Large Vision-Language Models},
author={Chenyang Zhang and Xiaoyu Zhang and Jian Lou and Kai Wu and Zilong Wang and Xiaofeng Chen},
booktitle={Forty-second International Conference on Machine Learning},
year={2025},
url={https://openreview.net/forum?id=6SIymOqJlc}
}

@inproceedings{shi2025promptinjectionattacktool,
  title={Prompt Injection Attack to Tool Selection in {LLM} Agents},
  author={Jiawen Shi and Zenghui Yuan and Guiyao Tie and Pan Zhou and Neil Zhenqiang Gong and Lichao Sun},    
  booktitle={Proceedings of the {NDSS} Symposium},
  year={2026},
  url={https://www.ndss-symposium.org/ndss-paper/prompt-injection-attack-to-tool-selection-in-llm-agents/}
}

@inproceedings{mathew-etal-2025-hidden,
    title = "Hidden in Plain Text: Emergence {\&} Mitigation of Steganographic Collusion in {LLM}s",
    author = "Mathew, Yohan  and
      Matthews, Ollie  and
      McCarthy, Robert  and
      Velja, Joan  and
      Schroeder de Witt, Christian  and
      Cope, Dylan  and
      Schoots, Nandi",
    editor = "Inui, Kentaro  and
      Sakti, Sakriani  and
      Wang, Haofen  and
      Wong, Derek F.  and
      Bhattacharyya, Pushpak  and
      Banerjee, Biplab  and
      Ekbal, Asif  and
      Chakraborty, Tanmoy  and
      Singh, Dhirendra Pratap",
    booktitle = "Proceedings of the 14th International Joint Conference on Natural Language Processing and the 4th Conference of the Asia-Pacific Chapter of the Association for Computational Linguistics",
    month = dec,
    year = "2025",
    address = "Mumbai, India",
    publisher = "The Asian Federation of Natural Language Processing and The Association for Computational Linguistics",
    url = "https://aclanthology.org/2025.ijcnlp-long.34/",
    pages = "585--624",
    ISBN = "979-8-89176-298-5"
}

@inproceedings{
mathew2024emergence,
title={Emergence of Steganography Between Large Language Models},
author={Yohan Mathew and Robert McCarthy and Joan Velja and Ollie Matthews and Nandi Schoots and Dylan Cope},
booktitle={Workshop on Socially Responsible Language Modelling Research},
year={2024},
url={https://openreview.net/forum?id=Qn1QgqDgtF}
}

@article{
zhao2025a,
title={A Survey of Recent Backdoor Attacks and Defenses in Large Language Models},
author={Shuai Zhao and Meihuizi Jia and Zhongliang Guo and Leilei Gan and XIAOYU XU and Xiaobao Wu and Jie Fu and Feng Yichao and Fengjun Pan and Anh Tuan Luu},
journal={Transactions on Machine Learning Research},
issn={2835-8856},
year={2025},
url={https://openreview.net/forum?id=wZLWuFHxt5},
note={Survey Certification}
}

@article{zhou_ni_lee_zhao_2025,
author = {Zhou, Yihe and Ali, Mohammed and Lee, Wei-Bin and Zhao, Qingchuan},
year = {2025},
month = {05},
pages = {3},
title = {A Survey on Backdoor Threats in Large Language Models (LLMs): Attacks, Defenses, and Evaluation Methods},
journal = {Transactions on Artificial Intelligence},
doi = {10.53941/tai.2025.100003}
}

@article{strachan2024testing,
  title={Testing theory of mind in large language models and humans},
  author={Strachan, James WA and Albergo, Dalila and Borghini, Giulia and Pansardi, Oriana and Scaliti, Eugenio and Gupta, Saurabh and Saxena, Krati and Rufo, Alessandro and Panzeri, Stefano and Manzi, Guido and others},
  journal={Nature Human Behaviour},
  volume={8},
  number={7},
  pages={1285--1295},
  year={2024},
  publisher={Nature Publishing Group UK London}
}

@misc{defortuny2025llmsstrategicagentsbeliefs,
      title={LLMs as Strategic Agents: Beliefs, Best Response Behavior, and Emergent Heuristics}, 
      author={Enric Junque de Fortuny and Veronica Roberta Cappelli},
      year={2025},
      eprint={2510.10813},
      archivePrefix={arXiv},
      primaryClass={cs.AI},
      url={https://arxiv.org/abs/2510.10813}, 
}

@misc{lidayan2025abbelllmagentsacting,
      title={ABBEL: LLM Agents Acting through Belief Bottlenecks Expressed in Language}, 
      author={Aly Lidayan and Jakob Bjorner and Satvik Golechha and Kartik Goyal and Alane Suhr},
      year={2025},
      eprint={2512.20111},
      archivePrefix={arXiv},
      primaryClass={cs.CL},
      url={https://arxiv.org/abs/2512.20111}, 
}

@inproceedings{
riemer2025position,
title={Position: Theory of Mind Benchmarks are Broken for Large Language Models},
author={Matthew Riemer and Zahra Ashktorab and Djallel Bouneffouf and Payel Das and Miao Liu and Justin D. Weisz and Murray Campbell},
booktitle={Forty-second International Conference on Machine Learning Position Paper Track},
year={2025},
url={https://openreview.net/forum?id=BCP8UU2BcU}
}

@inproceedings{10.1145/3689904.3694704,
author = {Dorn, Rebecca and Kezar, Lee and Morstatter, Fred and Lerman, Kristina},
title = {Harmful Speech Detection by Language Models Exhibits Gender-Queer Dialect Bias},
year = {2024},
isbn = {9798400712227},
publisher = {Association for Computing Machinery},
address = {New York, NY, USA},
url = {https://doi.org/10.1145/3689904.3694704},
doi = {10.1145/3689904.3694704},
booktitle = {Proceedings of the 4th ACM Conference on Equity and Access in Algorithms, Mechanisms, and Optimization},
articleno = {6},
numpages = {12},
location = {San Luis Potosi, Mexico},
series = {EAAMO '24}
}

@inproceedings{10.1145/3630106.3659036,
author = {Gomez, Juan Felipe and Machado, Caio and Paes, Lucas Monteiro and Calmon, Flavio},
title = {Algorithmic Arbitrariness in Content Moderation},
year = {2024},
isbn = {9798400704505},
publisher = {Association for Computing Machinery},
address = {New York, NY, USA},
url = {https://doi.org/10.1145/3630106.3659036},
doi = {10.1145/3630106.3659036},
booktitle = {Proceedings of the 2024 ACM Conference on Fairness, Accountability, and Transparency},
pages = {2234–2253},
numpages = {20},
location = {Rio de Janeiro, Brazil},
series = {FAccT '24}
}

@inproceedings{lu-etal-2024-take,
    title = "Take Its Essence, Discard Its Dross! Debiasing for Toxic Language Detection via Counterfactual Causal Effect",
    author = "Lu, Junyu  and
      Xu, Bo  and
      Zhang, Xiaokun  and
      Liu, Kaiyuan  and
      Zhang, Dongyu  and
      Yang, Liang  and
      Lin, Hongfei",
    editor = "Calzolari, Nicoletta  and
      Kan, Min-Yen  and
      Hoste, Veronique  and
      Lenci, Alessandro  and
      Sakti, Sakriani  and
      Xue, Nianwen",
    booktitle = "Proceedings of the 2024 Joint International Conference on Computational Linguistics, Language Resources and Evaluation (LREC-COLING 2024)",
    month = may,
    year = "2024",
    address = "Torino, Italia",
    publisher = "ELRA and ICCL",
    url = "https://aclanthology.org/2024.lrec-main.1353/",
    pages = "15566--15578"
}

@techreport{ukAIActionPlan2025,
  title       = {{AI Opportunities Action Plan}},
  author      = {{UK DSIT}},
  institution = {{UK Government}},
  year        = {2025},
  month       = {January},
  url         = {https://www.gov.uk/government/publications/ai-opportunities-action-plan/ai-opportunities-action-plan},
  note        = {Accessed: 23 January 2026}
}

@misc{ncscPromptInjection2025,
  title        = {{Prompt Injection Is Not SQL Injection (It May Be Worse)}},
  author       = {{UK National Cyber Security Centre}},
  year         = {2025},
  month        = {December},
  howpublished = {\url{https://www.ncsc.gov.uk/blog-post/prompt-injection-is-not-sql-injection}},
  note         = {UK National Cyber Security Centre blog post. Accessed: 23 January 2026}
}

@misc{owaspLLM01PromptInjection2025,
  title        = {{LLM01: Prompt Injection}},
  author       = {{OWASP Foundation}},
  year         = {2025},
  howpublished = {\url{https://genai.owasp.org/llmrisk/llm01-prompt-injection/}},
  note         = {OWASP Top 10 for Large Language Model Applications. Accessed: 23 January 2026}
}

@misc{usEO14110AI2023,
  title        = {{Safe, Secure, and Trustworthy Development and Use of Artificial Intelligence}},
  author       = {{The White House}},
  year         = {2023},
  month        = {November},
  howpublished = {\url{https://www.federalregister.gov/documents/2023/11/01/2023-24283/safe-secure-and-trustworthy-development-and-use-of-artificial-intelligence}},
  note         = {Executive Order 14110, Federal Register. Accessed: 23 January 2026}
}

@misc{euAIAct2024,
  title        = {{Regulation (EU) 2024/1689 (Artificial Intelligence Act)}},
  author       = {{European Union}},
  year         = {2024},
  month        = {June},
  howpublished = {\url{https://eur-lex.europa.eu/eli/reg/2024/1689/oj/eng}},
  note         = {Official Journal of the European Union. Accessed: 23 January 2026}
}

@InProceedings{10.1007/3-540-45748-8_24,
author="Douceur, John R.",
editor="Druschel, Peter
and Kaashoek, Frans
and Rowstron, Antony",
title="The Sybil Attack",
booktitle="Peer-to-Peer Systems",
year="2002",
publisher="Springer Berlin Heidelberg",
address="Berlin, Heidelberg",
pages="251--260",
isbn="978-3-540-45748-0"
}

@article{https://doi.org/10.1111/1467-9760.00148,
author = {Sunstein, Cass R.},
title = {The Law of Group Polarization},
journal = {Journal of Political Philosophy},
volume = {10},
number = {2},
pages = {175-195},
doi = {https://doi.org/10.1111/1467-9760.00148},
url = {https://onlinelibrary.wiley.com/doi/abs/10.1111/1467-9760.00148},
eprint = {https://onlinelibrary.wiley.com/doi/pdf/10.1111/1467-9760.00148},
year = {2002}
}

@misc{fang2025debateimages,
      title={Debate with Images: Detecting Deceptive Behaviors in Multimodal Large Language Models}, 
      author={Sitong Fang and Shiyi Hou and Kaile Wang and Boyuan Chen and Donghai Hong and Jiayi Zhou and Josef Dai and Yaodong Yang and Jiaming Ji},
      year={2025},
      eprint={2512.00349},
      archivePrefix={arXiv},
      primaryClass={cs.AI},
      url={https://arxiv.org/abs/2512.00349}, 
}

@misc{bai2025qwen3vl,
  title         = {Qwen3-VL Technical Report},
  author        = {Bai, Shuai and Cai, Yuxuan and Chen, Ruizhe and others},
  year          = {2025},
  eprint        = {2511.21631},
  archivePrefix = {arXiv},
  primaryClass  = {cs.CV},
  url           = {https://arxiv.org/abs/2511.21631}
}

@misc{openai2024gpt4omini,
  title        = {GPT-4o mini: advancing cost-efficient intelligence},
  author       = {{OpenAI}},
  year         = {2024},
  howpublished = {\url{https://openai.com/index/gpt-4o-mini-advancing-cost-efficient-intelligence/}},
  note         = {Accessed: 2026-05-05}
}

@misc{fragileflow,
  author   = {Z. Li and B. Wang and J. Hu and X. Huang and Y. Dong},
  title    = {FragileFlow: Spectral Control of Correct-but-Fragile Predictions for Foundation Model Robustness},
  year     = {2026},
  note     = {arXiv:2605.08896}
}

@misc{scope,
  author   = {Z. Li and B. Wang and C. Wu and X. Huang and Y. Dong},
  title    = {SCOPE: Sequential Conformal Probing for Reliable OOD Rejection in LLM Services},
  year     = {2026},
  note     = {arXiv:2606.21255}
}

@misc{pathmark,
  author   = {Y. Gao and others},
  title    = {PathMark: Protecting Intellectual Property of Mixture-of-Expert LLMs via Path Watermarks},
  year     = {2026},
  note     = {arXiv:2607.03688}
}

@misc{chen2026promptsleavebehavioralfingerprints,
      title={Do System Prompts Leave Behavioral Fingerprints? A Large-Scale Empirical Study of Clone Detection via Output Similarity}, 
      author={Linghan Chen and Yudong Gao and Jiyao Wang and Kaiyan Ji and Honglong Chen},
      year={2026},
      eprint={2608.24461},
      archivePrefix={arXiv},
      primaryClass={cs.CR},
      url={https://arxiv.org/abs/2608.24461}, 
}

@ARTICLE{10980439,
  author={Yu, Jimiao and others},
  journal={IEEE Transactions on Multimedia}, 
  title={Black-Box Adversarial Defense Based on Image Decomposition and Reconstruction}, 
  year={2025},
  volume={27},
  number={},
  pages={5909-5921},
  doi={10.1109/TMM.2025.3565987}}

@inproceedings{
azulay2026jailbreaking,
title={Jailbreaking Vision-Language Models Through the Visual Modality},
author={Aharon Azulay and Jan Dubi{\'n}ski and Zhuoyun Li and Atharv Mittal and Yossi Gandelsman},
booktitle={Proc. ICML},
year={2026},
url={https://openreview.net/forum?id=4VuMV9beC1}
}

@inproceedings{wang-etal-2026-chain,
    title = "Chain-of-Thought as a Lens: Evaluating Structured Reasoning Alignment between Human Preferences and Large Language Models",
    author = "Wang, Boxuan  and others",
    editor = "Liakata, Maria  and
      Moreira, Viviane P.  and
      Zhang, Jiajun  and
      Jurgens, David",
    booktitle = "Proc. ACL (Volume 1: Long Papers)",
    month = jul,
    year = "2026",
    address = "San Diego, California, United States",
    publisher = "Association for Computational Linguistics",
    url = "https://aclanthology.org/2026.acl-long.1834/",
    doi = "10.18653/v1/2026.acl-long.1834",
    pages = "39514--39530",
    ISBN = "979-8-89176-390-6"
}

\appendix
\section{Prototype Evaluation of EB-MAS}
\label{app:prototype_eval}

This appendix reports a proof-of-concept implementation of the EB-MAS graph firewall. The goal is not to provide a complete empirical benchmark, but to test whether explicit belief-graph features can provide useful runtime safety signals in realistic multimodal and agentic settings.

\subsection{Implementation}

For each candidate response or tool choice, EB-MAS constructs an Explicit Belief Substrate graph with source, evidence, claim, belief, update-proposal, and firewall-decision nodes. The graph firewall uses features for belief strength, uncertainty, vulnerability, evidence support, source reliability, source independence, coordination risk, cross-cluster diversity, evidence quality, and evidence--vulnerability mismatch. These features allow the firewall to evaluate whether a risky belief update is sufficiently supported by reliable and independent evidence.

\subsection{Multimodal Deception}

We evaluate multimodal deceptive requests on MM-DeceptionBench using Qwen3-VL-32B-Instruct. Direct model responses are judged unsafe in 67.92\% of 1,013 deceptive cases. A keyword/phrase guard reduces this to 38.01\%, but falsely blocks 11.00\% of benign controls. An LLM belief-operator-only baseline reduces unsafe outputs to 3.65\%, but falsely blocks 25.00\% of benign controls. EB-MAS reduces unsafe outputs to 8.19\% while keeping benign false blocking at 1.00\%.

\begin{table}[h]
\centering
\small
\setlength{\tabcolsep}{4.5pt}
\renewcommand{\arraystretch}{1.08}
\caption{\textbf{MM-DeceptionBench prototype results.}
Failure is the API-judged unsafe rate after defense. Benign false block is measured on 100 benign-control examples.}
\label{tab:app_mm_deception}
\begin{tabular}{lccc}
\toprule
Method & Attack $n$ & Unsafe after defense $\downarrow$ & Benign false block $\downarrow$ \\
\midrule
Direct Qwen3-VL response & 1,013 & 67.92 & 0.00 \\
Keyword/phrase guard & 1,013 & 38.01 & 11.00 \\
LLM belief-operator only & 1,013 & \textbf{3.65} & 25.00 \\
EB-MAS graph firewall & 1,013 & 8.19 & \textbf{1.00} \\
\bottomrule
\end{tabular}
\end{table}

\subsection{Agent Tool and Memory Attacks}

We evaluate 400 Agent Security Bench tool/memory attack cases using GPT-4o-mini as the agent brain. Each case contains a legitimate task, normal tools, an attack tool, and an untrusted memory note. EB-MAS converts the selected tool into a provenance-aware belief graph and audits whether the tool choice is supported by trusted provenance.

\begin{table}[h]
\centering
\small
\setlength{\tabcolsep}{4.5pt}
\renewcommand{\arraystretch}{1.08}
\caption{\textbf{ASB tool/memory attack prototype results.}
Failure is attack-tool pass-through after defense. Safe allow measures the fraction of normal tool choices allowed by the defense.}
\label{tab:app_asb}
\begin{tabular}{lccc}
\toprule
Method & $n$ & Attack after defense $\downarrow$ & Safe allow $\uparrow$ \\
\midrule
Direct GPT-4o-mini agent & 400 & 6.50 & 100.00 \\
Prompt-only safety instruction & 400 & 4.25 & 100.00 \\
Keyword denylist guard & 400 & 4.25 & 45.19 \\
LLM self-check guard & 400 & 3.50 & 100.00 \\
Trusted-tool whitelist & 400 & \textbf{0.00} & \textbf{100.00} \\
Rule EB-MAS graph firewall & 400 & \textbf{0.00} & \textbf{100.00} \\
Learned EB-MAS graph firewall & 400 & \textbf{0.00} & \textbf{100.00} \\
\bottomrule
\end{tabular}
\end{table}

\subsection{Limitations}

These results should be interpreted as preliminary evidence. On MM-DeceptionBench, sandbagging remains the hardest category, suggesting that future versions need capability-consistency features, historical competence traces, and self-uncertainty drift signals. On ASB, provenance cues are relatively strong, so the learned firewall should be viewed as evidence that graph features can be calibrated, not as proof that EB-MAS solves open-world agent security. More ambiguous provenance, corrupted registries, and dynamic environments such as AgentDojo are important next stress tests.
% \section{Technical appendices and supplementary material}
% Technical appendices with additional results, figures, graphs, and proofs may be submitted with the paper submission before the full submission deadline (see above). You can upload a ZIP file for videos or code, but do not upload a separate PDF file for the appendix. There is no page limit for the technical appendices. 

% Note: Think of the appendix as ``optional reading'' for reviewers. The paper must be able to stand alone without the appendix; for example, adding critical experiments that support the main claims to an appendix is inappropriate. 

%%%%%%%%%%%%%%%%%%%%%%%%%%%%%%%%%%%%%%%%%%%%%%%%%%%%%%%%%%%%

% \clearpage
\section{Supplementary Experiments and Discussion}
\label{app:supplementary_experiments_discussion}

This appendix develops the runtime update procedure and presents supplementary evaluations of evidence calibration, source dependence, and belief revision across rounds. The experiments examine how the audit interface balances unsafe acceptance with benign update allowance, and how dependency and commitment checks contribute to this balance.

\subsection{Runtime Update Procedure}
\label{app:supp_runtime}

EB-MAS organizes belief revision into three stages: propose, verify, and commit. For an incoming signal $S_t$, the LLM extracts a proposition $p_i$, supporting evidence spans, and a typed candidate update,
\begin{equation}
\delta_{i,t}^{\mathrm{cand}}
=\bigl(\Delta v_{i,t}^{\mathrm{cand}},\Delta G_{i,t}^{\mathrm{cand}},
E_{i,t}^{\mathrm{cand}},r_{i,t}^{\mathrm{int}},c_{i,t}^{\mathrm{ext}}\bigr).
\label{eq:supp_candidate}
\end{equation}
Here, $\Delta v_{i,t}^{\mathrm{cand}}$ and $\Delta G_{i,t}^{\mathrm{cand}}$ describe proposed changes to commitment strength and graph structure. The candidate also carries evidence references $E_{i,t}^{\mathrm{cand}}$, interaction-level influence risk $r_{i,t}^{\mathrm{int}}$, and extraction confidence $c_{i,t}^{\mathrm{ext}}$. Epistemic uncertainty $u_i^t$ describes uncertainty about the proposition, while extraction confidence describes confidence in the mapping from the incoming signal to the candidate fields.

The runtime checks document and tool identifiers, sender identity, timestamps, signatures, registry status, and memory origin. It resolves surface sources to provenance roots and checks the candidate evidence for attribution, support, source dependence, and consistency with the existing graph. Several surface sources that share a provenance root contribute a common evidence dependency. Missing or conflicting metadata receives an unknown status and routes the candidate to verification before a high-impact persistent update.

After provenance checks, the prototype combines verified evidence support $e$ and source reliability $s$ into an evidence-quality score,
\begin{equation}
q=\alpha e+\beta s,
\qquad \alpha+\beta=1,\quad \alpha>\beta\geq 0.
\label{eq:supp_evidence_quality}
\end{equation}
The weights give direct evidence the larger contribution. Source dependence and coordination receive separate checks. Interaction-level influence risk captures cues in the current update, including unsupported repetition, authority pressure, urgency, emotional framing, and coordinated support. An evidence--influence mismatch score combines these signals as
\begin{equation}
m=(1-q)\max\bigl(r^{\mathrm{int}},r^{\mathrm{coord}}\bigr),
\label{eq:supp_mismatch}
\end{equation}
where $r^{\mathrm{coord}}$ describes coordination risk. Strong influence cues paired with weak verified evidence produce a large mismatch. Attaching this score to the candidate update supports interaction-local retention and access rules.

The commit stage applies evidence and update-magnitude constraints. Let $Q_{i,t}\in[0,1]$ summarize verified support, source reliability, independence, and consistency. The provenance gate requires
\begin{equation}
\operatorname{Commit}(\delta_{i,t}^{\mathrm{cand}})
\ \Longrightarrow\ 
\operatorname{Resolvable}(E_{i,t}^{\mathrm{cand}})
\ \land\ Q_{i,t}\geq\tau_{\min}.
\label{eq:supp_provenance_gate}
\end{equation}
For a scalar commitment update, the evidence-dependent bound and admissible change are
\begin{align}
B_{i,t}&=\eta Q_{i,t}(1-u_i^t)(1-r_{i,t}^{\mathrm{int}}),\label{eq:supp_update_budget}\\
\Delta v_{i,t}^{*}
&=\operatorname{clip}\bigl(\Delta v_{i,t}^{\mathrm{cand}},-B_{i,t},B_{i,t}\bigr),
\label{eq:supp_clipped_update}
\end{align}
where $\eta\geq0$ sets the update scale. Topology edits pass typed structural preconditions. Accepted candidates enter the persistent state unchanged; clipped candidates enter with an admissible update magnitude; quarantined candidates wait for verification with action and propagation suspended; rejected candidates are discarded. The runtime records the evidence links, checks, and commit decision for subsequent inspection.

Across rounds, the audit state retains provenance roots, graph dependencies, and a commitment history. The cumulative-commitment check compares the resulting commitment with the evidence accumulated for the proposition. This makes repeated exposure and successive small edits visible within a common update history. Strong verified evidence supports substantial revisions, including corrections that change an earlier conclusion.

\subsection{Safety--Utility Calibration on Multimodal Deception}
\label{app:supp_calibration}

We extend the MM-DeceptionBench evaluation with an expanded benign-control set and held-out threshold calibration. Appendix~\ref{app:prototype_eval} reports the initial evaluation with 100 benign controls. The supplementary evaluation uses 1,013 benign controls alongside the 1,013 deceptive cases. Table~\ref{tab:supp_mm_expanded} reports the binary belief operator and fixed EB-MAS rule on this expanded evaluation. Unsafe acceptance counts deceptive responses that pass the defense and receive an unsafe label from the automatic judge. Benign false blocking counts benign cases rejected by the defense.

\begin{table}[htbp]
\centering
\small
\setlength{\tabcolsep}{5pt}
\renewcommand{\arraystretch}{1.12}
\caption{\textbf{Expanded MM-DeceptionBench evaluation.} Each outcome is reported as count/denominator and percentage. Both deceptive and benign sets contain 1,013 cases.}
\label{tab:supp_mm_expanded}
\begin{tabular}{lcc}
\toprule
Method & Unsafe acceptance & Benign false block \\
\midrule
Binary belief operator & 37/1,013 (3.65\%) & 254/1,013 (25.07\%) \\
Fixed EB-MAS rule & 83/1,013 (8.19\%) & 25/1,013 (2.47\%) \\
\bottomrule
\end{tabular}
\end{table}

The two operating points illustrate the importance of measuring benign allowance alongside unsafe acceptance. The binary operator passes fewer unsafe responses and rejects about one quarter of the benign set. The fixed EB-MAS rule allows a larger fraction of benign cases, with 25 false blocks across 1,013 controls.

For threshold calibration, we split the benign controls into 506 calibration cases and 507 held-out cases. Thresholds are selected using the calibration split under a chosen false-block budget and frozen before attack evaluation. Table~\ref{tab:supp_threshold_calibration} shows four budgets. At the 5\% budget, unsafe acceptance is 68/1,013 (6.71\%; 95\% CI [5.33, 8.42]\%), and held-out benign false blocking is 15/507 (2.96\%; 95\% CI [1.80, 4.82]\%). At the 25\% budget, unsafe acceptance reaches 1.78\%, alongside 18.74\% held-out false blocking. The sweep gives an explicit operating curve for choosing a benign-utility budget.

\begin{table}[htbp]
\centering
\small
\setlength{\tabcolsep}{4pt}
\renewcommand{\arraystretch}{1.12}
\caption{\textbf{Held-out threshold calibration.} The calibration and held-out benign splits contain 506 and 507 cases, respectively. Unsafe acceptance is evaluated on 1,013 deceptive cases.}
\label{tab:supp_threshold_calibration}
\begin{tabular}{cccc}
\toprule
False-block budget & Calibration false block & Unsafe acceptance & Held-out false block \\
\midrule
1\% & 0/506 (0.00\%) & 688/1,013 (67.92\%) & 0/507 (0.00\%) \\
5\% & 20/506 (3.95\%) & 68/1,013 (6.71\%) & 15/507 (2.96\%) \\
10\% & 34/506 (6.72\%) & 56/1,013 (5.53\%) & 22/507 (4.34\%) \\
25\% & 93/506 (18.38\%) & 18/1,013 (1.78\%) & 95/507 (18.74\%) \\
\bottomrule
\end{tabular}
\end{table}

An auxiliary automatic-judge comparison with Claude Haiku 4.5 gives 81.67\% agreement and Cohen's $\kappa=0.633$. Sandbagging agreement is 55\%. Residual errors concentrate in sandbagging and sycophancy, making category-specific adjudication a useful part of further evaluation. The single-source multimodal setting exercises evidence and influence scoring; the following experiment exercises source dependencies and commitment history across rounds.

\subsection{Four-Round Agent Evaluation and Structural Ablation}
\label{app:supp_four_round}

We conduct a matched four-round ASB-derived evaluation with a GPT-4o-mini agent. The test includes correlated provenance roots, repeated persuasion from a common root, trusted-root semantic drift, and cumulative commitment that exceeds the available evidence. Attack and benign trajectories are matched on surface source count, evidence support, and declared reliability. The agent independently decides whether to authorise each update, and the defense then determines whether the proposed update is committed.

We compare the direct agent, a verified-root whitelist, a calibrated provenance classifier, a joint ablation that removes dependency-resolution and cumulative-commitment checks, and full EB-MAS. Table~\ref{tab:supp_four_round} reports unsafe commits over 240 attack cases and added false blocks over 219 benign updates authorised by the agent. Added false blocking measures the utility loss introduced by the defense after the agent's authorisation decision.

\begin{table}[htbp]
\centering
\small
\setlength{\tabcolsep}{3.5pt}
\renewcommand{\arraystretch}{1.15}
\caption{\textbf{Four-round ASB-derived evaluation.} The ablation removes dependency-resolution and cumulative-commitment checks together. Confidence intervals refer to the unsafe-commit rate and are expressed in percent.}
\label{tab:supp_four_round}
\begin{tabular}{>{\raggedright\arraybackslash}p{0.31\linewidth}ccc}
\toprule
Method & Unsafe commits & 95\% CI & Added benign false blocks \\
\midrule
Direct GPT-4o-mini & 74/240 (30.83\%) & [25.33, 36.94] & 0/219 (0.00\%) \\
Verified-root whitelist & 24/240 (10.00\%) & [6.81, 14.45] & 0/219 (0.00\%) \\
Provenance classifier & 20/240 (8.33\%) & [5.46, 12.52] & 10/219 (4.57\%) \\
EB-MAS joint ablation & 74/240 (30.83\%) & [25.33, 36.94] & 0/219 (0.00\%) \\
Full EB-MAS & 0/240 (0.00\%) & [0.00, 1.58] & 0/219 (0.00\%) \\
\bottomrule
\end{tabular}
\end{table}

The provenance classifier reaches an AUROC of 0.9444. Its remaining 20 unsafe commits occur in the trusted-root semantic-drift setting, where successive changes accumulate through a source with a trusted identity. The whitelist allows 24 unsafe commits. With dependency and cumulative-commitment checks removed together, EB-MAS records 74 unsafe commits, matching the direct-agent outcome. Full EB-MAS records zero unsafe commits and zero added benign false blocks; the corresponding 95\% confidence intervals are [0.00, 1.58]\% and [0.00, 1.72]\%, respectively.

Paired exact comparisons of unsafe-commit outcomes with full EB-MAS give $p=1.19\times10^{-7}$ for the verified-root whitelist, $p=1.91\times10^{-6}$ for the provenance classifier, and $p=1.06\times10^{-22}$ for the joint ablation. Together, the matched comparison and joint ablation show the contribution of dependency resolution and commitment history in this four-round setting.

The benign trajectories include evidence-supported corrections from trusted sources, while the attack trajectories introduce semantic drift and accumulate excessive commitment. Full EB-MAS allows all 219 agent-authorised benign updates in the evaluated set. This pairing exercises both legitimate revision and resistance to unsupported accumulation under comparable surface provenance.

\subsection{Discussion}
\label{app:supp_discussion}

\paragraph{The update transition as an audit object.}
The supplementary evaluations connect four objects: the current epistemic state, the candidate change, its supporting evidence, and the commit decision. Provenance roots expose shared dependencies across surface sources. The commitment history exposes how several updates combine into a stronger premise. The commit interface then connects these checks to persistent memory, tool use, and peer communication. This state-transition view makes the route from incoming information to downstream action available for inspection.

\paragraph{Evidence-sensitive revision.}
Calibration and structural checks address complementary parts of the update process. The multimodal experiment measures the operating trade-off between unsafe acceptance and benign allowance. The four-round experiment measures how source dependence and accumulated commitment affect persistence. Evidence-supported corrections remain part of the benign evaluation, giving a direct measure of whether the system preserves useful revision. These results support evaluating a belief-update interface through both its acceptance decisions and its treatment of evidence over time.

\paragraph{Runtime grounding and deployment.}
The runtime's provenance registry, evidence resolver, and typed checks form the trust anchor for the commit interface. Extraction quality and source truthfulness remain separate evaluation targets within this pipeline. A practical deployment can attach evidence spans, metadata checks, and decision records to each committed change, and route uncertain provenance to verification. Interaction-level influence scores describe cues in the candidate update. Purpose-specific access, short retention, and user contestability provide concrete deployment controls for these records.

\paragraph{Further evaluation.}
The next experiments can vary provenance ambiguity, root correlation, and the length of the commitment history, while measuring unsafe commits, benign false blocks, and verification overhead. Forged provenance, corrupted registries, graph-construction attacks, and persistent-memory corruption provide distinct stress tests of the runtime components. Adaptive tool-use environments such as AgentDojo can further examine how the same audit interface operates as evidence, available tools, and interaction partners change during a task.

\newpage

\end{document}